\documentclass{article}

\usepackage[final]{neurips_2025}

\usepackage[utf8]{inputenc} %
\usepackage[T1]{fontenc}    %
\usepackage{hyperref}       %
\usepackage{url}            %
\usepackage{booktabs}       %
\usepackage{amsfonts}       %
\usepackage{nicefrac}       %
\usepackage{microtype}      %
\usepackage{xcolor}         %

\usepackage{amsmath}
\usepackage{cleveref}

\newcommand{\authnote}[3]{\textcolor{#3}{[{\footnotesize {\bf #1:} { {#2}}}]}}
\renewcommand{\authnote}[3]{}  %

\newcommand{\R}{\mathbb{R}}

\newcommand{\Nsamples}{N_\mathrm{sam}}
\newcommand{\Nsec}{N_\mathrm{sec}}

\newcommand{\MASK}{\mathtt{[MASK]}}

\usepackage{tcolorbox} %
\usepackage{tasks} %

\title{WhAM: Towards A Translative Model of Sperm Whale Vocalization}

\author{%
  Orr Paradise$^{1,2}$ \And
  Pranav Muralikrishnan$^{*1}$ \And
  Liangyuan Chen$^{*1}$ \And
  Hugo Flores García$^{3}$ \And
  Bryan Pardo$^{3}$ \And
  Roee Diamant$^{4,2}$ \And
  David F. Gruber$^{5,2}$ \And
  Shane Gero$^{6,2}$ \And
  Shafi Goldwasser$^{1,2}$ \\
  \\
  $^1$UC Berkeley \quad
  $^2$Project CETI \quad
  $^3$Northwestern University \quad
  $^4$Haifa University \\
  $^5$City University of New York \quad
  $^6$Carleton University \\
  $^*$Equal contribution
}

\begin{document}

\maketitle

\begin{abstract}
  Sperm whales communicate in short sequences of clicks known as codas. We present WhAM (Whale Acoustics Model), the first transformer-based model capable of generating synthetic sperm whale codas from any audio prompt. WhAM is built by finetuning VampNet, a masked acoustic token model pretrained on musical audio, using 10k coda recordings collected over the past two decades. Through iterative masked token prediction, WhAM generates high-fidelity synthetic codas that preserve key acoustic features of the source recordings. We evaluate WhAM's synthetic codas using Fr\'{e}chet Audio Distance and through perceptual studies with expert marine biologists. On downstream classification tasks including rhythm, social unit, and vowel classification, WhAM's learned representations achieve strong performance, despite being trained for generation rather than classification. Our code is available at \url{https://github.com/Project-CETI/wham}
\end{abstract}

\section{Introduction}
Understanding the communication of sperm whales (\emph{Physeter macrocephalus}) is among the most fascinating questions in animal behavioral studies.

Sperm whales communicate using \emph{codas}---short sequences of clicks that vary in number, rhythm, and tempo \citep{WatkinsS77,WeilgartW93,SharmaGPGRTA24}. They live in stable, female-led social units that form larger vocal clans based on \emph{dialect} \citep{RendellW03}. That is, the dialect of a social unit determines its clan, with social units associating exclusively with other units from their clan \citep{GeroBWM16}. Furthermore, dialects are believed to be learned socially rather than inherited genetically \citep{CantorW15,RendellMDBW12}.

The complexity of these learned vocal patterns has motivated new computational approaches to understanding codas and their functionality. \citet{LeitaoEtAl24} modeled codas as (variable-length) Markov chains, revealing new patterns of inter-clan social learning. \citet{Vowels} study vowel-like spectral properties of codas, which were initially suggested by interpreting the codebook of a Generative Adversarial Network (GAN). Most recently, \citet{WhaleLM} trained a transformer on click timings (inter-click intervals), which is able to predict codas in an exchange based on long-term dependencies, as well as future diving behavior. These studies collectively highlight how machine learning---particularly transformer architectures---can decode patterns imperceptible to traditional methods.

Transformers \citep{VaswaniSPUJGKP17} originated in natural language translation, where they revolutionized the field by enabling high-quality, context-aware machine translation.
Whereas transformers have since become ubiquitous across machine learning (e.g. \citealt{ChenLRLGLASM21,KhanNHZKS22,MoussadRB23}), in this work we propose again to use transformers towards translation---of animal communication.

While transformers have been used in settings where parallel data is nonexistent \citep{ConneauL19} and for translation from audio \citep{KanoSN021a}, applying these advances to animal communication presents deep challenges. Even merely defining the problem has been the subject of studies spanning theoretical computer science \citep{GoldwasserGKP23}, biology \citep{YovelR23, AmphearisEtAl23}, linguistics \citep{BerwickC16, AmphearisST22}, and philosophy \citep{SuzukiWG20, HobaiterGB22}.

\begin{figure*}[t]
    \centering
    \includegraphics[width=\linewidth]{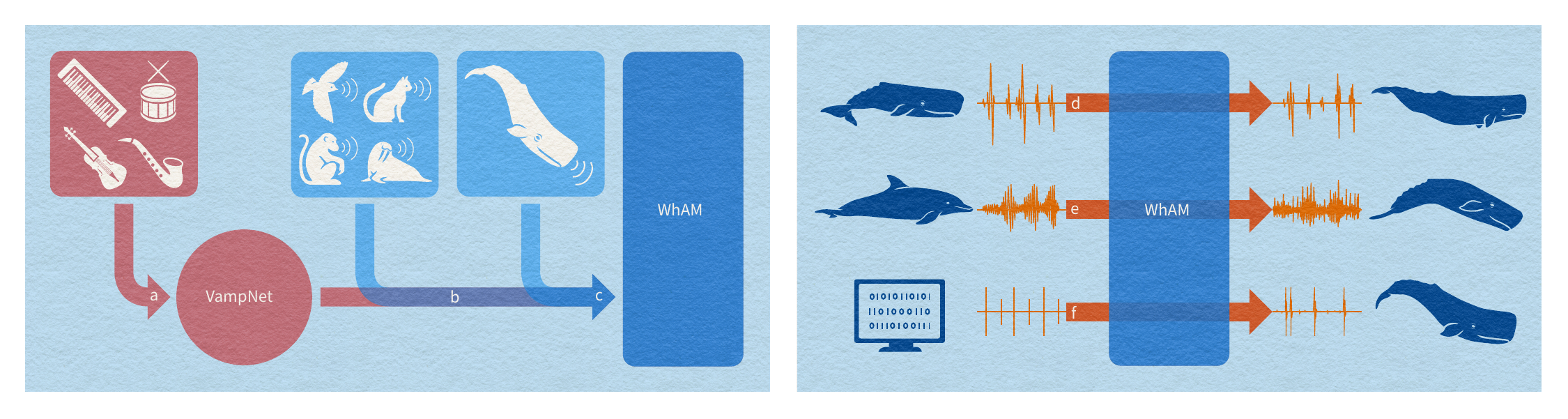}
    \caption{
        Left: WhAM is trained by finetuning VampNet \citep{GarciaSKP23}, an audio-to-audio transformer pretrained on a large music dataset (a). Namely, we perform \textbf{domain adaptation} (b) on animal vocalizations followed by \textbf{species-specific finetuning} (c) on a novel sperm whale coda dataset.
        Right: WhAM synthesizes \textbf{context-aware variations} (d) of input codas and \textbf{acoustically translates} (e) natural and (f) artificial audio into coda-like audio. Illustration \textcopyright{} Alex Boersma 2025.
    }
    \label{fig:main}
\end{figure*}
Existing approaches to modeling sperm whale codas have made significant advances. \citet{BermantBWGG19} developed effective methods for coda detection and classification, while generative models based on GANs \citep{BegusLG23, Whales24} have shown the potential for synthesizing coda-like audio. The aforementioned timing-based analyses of \citet{LeitaoEtAl24,WhaleLM} have yielded new insight into the social and behavioral aspects of sperm whale communication.

Our work will address challenges left open by these works: While GAN-based models can generate coda-like audio \citep{BegusLG23,Whales24}, they cannot easily condition on a given context. Timing-based approaches \citep{LeitaoEtAl24,WhaleLM} capture important temporal patterns but may miss features only present in the raw audio, such as the recently discovered vowel-like spectral patterns \citep{Vowels}. Moreover, current methods train separate models for classification \citep{BermantBWGG19} and generation, despite the intuition that a model capable of realistic generation should also learn representations useful for classification. Lastly, none of these tackled the issue of \emph{translating across acoustic domains}.

To address these challenges, we introduce the Whale Acoustics Model (WhAM, \Cref{fig:main}), a new approach to modeling sperm whale codas that unifies three capabilities:
\begin{itemize}
\item \textbf{Acoustic translation:}\footnote{We emphasize that translation is in the acoustic sense; semantic translation remains a distinct and more ambitious goal.} WhAM can translate an audio prompt (e.g. other animal vocalizations or even noise) into the acoustic style of sperm whale codas, acting as a form of cross-domain style transfer.
\item \textbf{Generation:} WhAM can generate novel ``pseudocodas'' that are perceptually similar to real codas, as evaluated by expert listeners.
\item \textbf{Classification:} WhAM's learned representations are useful for a range of classification tasks, including rhythm type \citep{SharmaGPGRTA24}, social unit classification \citep{Best79,ChristalW01,GeroWR16}, and the recently discovered vowel-like spectral patterns of \citet{Vowels}---despite being trained primarily for generation.
\end{itemize}

\paragraph{Contributions.} This paper presents the first unified model of sperm whale codas capable of acoustic translation, generation, and classification. Notably, WhAM demonstrates that meaningful bioacoustic features emerge from purely generative training, aligning with recent work on self-supervised (non-generative) modeling of animal vocalizations \citep{Hagiwara23}.

WhAM serves as a proof of concept, applying advances in neural audio modeling to bioacoustics in a novel and unifying way. To facilitate further research, we will release the model and its training and evaluation code. Remarkably, WhAM achieves strong results after just five days of training on a single GPU. While the dataset is small compared to those used for large audio models \citep{AudioLM23,MusicLM23}, our results suggest that scaling up could yield even greater improvements.

Finally, WhAM was developed in close collaboration with marine biologists and underwater acousticians with domain expertise in sperm whale vocalizations. The model was evaluated through perceptual studies conducted by an interdisciplinary team of specialists. To our knowledge, this is the first paper to evaluate the perception of experts on synthetically generated codas, igniting a crucial discourse for validating the utility of generative models in bioacoustics research. Code, model, and data are available at \url{https://github.com/Project-CETI/wham}

\paragraph{Outline.} \Cref{sec:related} reviews related work. \Cref{sec:methods} details our methodological framework. \Cref{sec:results} presents experimental results and expert analysis. \Cref{sec:future} discusses future work.

\section{Related work}\label{sec:related}

\paragraph{Audio Generation.}
The vast majority of studies on deep generative audio models focus on human speech or music (e.g. \citealt{WaveNet16,MuseGan18,Jukebox20,LakhotiaEtAl21,MusicLM23}). Some works are dedicated to generating the vocalizations of animals (bioacoustics) such as birds \citep{Birds22,Birds24}, mice \citep{Mice23}, cetaceans \citep{Orcas22,Dolphin22a,Dolphin22b,Dolphin24}, and in particular sperm whales \citep{BegusLG23,Whales24}. However, to our knowledge, all techniques for bioacoustic generation are based on generative adversarial networks (GANs). 
Unlike our transformer-based WhAM, GANs do not allow for conditioning on context in the form of an audio prompt. We emphasize that WhAM enables translation of input sounds into the \emph{acoustic style} of sperm whale vocalizations, operating purely at the signal level. This is distinct from \emph{semantic} translation between communication systems, which remains a far more ambitious goal requiring a deep understanding of animal cognition and communication (e.g. \citealt{GoldwasserGKP23,YovelR23,AmphearisEtAl23}).

\paragraph{Animal Vocalization Modeling.}
Deep learning techniques have been applied towards other, non-generative, ends in bioacoustics research. Learned audio representations have been used for species recognition \cite{SpeciesRecog14,ForestSpecies14,MarineSpecies19,BirdNet21,BirdSurvey23} and automatic annotation (i.e., vocalization detection and classification) of bioacoustic data \cite{OrcaSpot19,DeepSqueak19,BermantBWGG19,MiceClass21}. AVES \citep{Hagiwara23} utilizes HuBERT's \citep{HuBERT21} self-supervised learning framework towards state-of-the-art performance in species classification and detection tasks. While AVES demonstrates the power of learned audio representations, its encoder-only architecture limits it to analysis tasks, contrasting with WhAM's generative capabilities. As we show in \Cref{sec:downstream}, while AVES outperforms WhAM on classification tasks as expected given its specialized design, WhAM still learns meaningful representations as a byproduct of its generative training, outperforming baseline approaches despite having a different primary objective.

\paragraph{Sperm whale communication.}
Understanding sperm whale communication has been a central challenge in marine biology for over six decades (\citealt{BackusS66,WatkinsS77,WhiteheadW91,AndreasEtAl22}; see also \Cref{apx:codas}). Recent computational approaches have focused on analyzing click timing patterns within codas and do not directly address the acoustic properties of individual clicks within codas \citep{SharmaGPGRTA24,LeitaoEtAl24,WhaleLM}. WhAM extends this computational trajectory by enabling systematic manipulation of click acoustics, potentially allowing a quantitative analysis of acoustic variations between clan dialects and investigation of features that make codas recognizable. While WhAM's synthetic codas may not yet match the quality needed for playback experiments, WhAM represents progress towards stimuli generation in a responsible behavioral study (\citealt{Playback1,Playback2,Playback3}; see also \Cref{sec:impact}).

\section{Methods}\label{sec:methods}

\subsection{Masked Acoustic Token Modeling with VampNet}\label{sec:VN}

VampNet \cite{GarciaSKP23} is an audio-to-audio generative model, pretrained on 797k music tracks from thousands of artists.  VampNet consists of three neural models: a tokenizer, a coarse-token model, and a coarse-to-fine model. For simplicity of presentation we will avoid the distinction between coarse and fine tokens, instead decomposing VampNet into an \emph{Acoustic Tokenizer} and a \emph{Masked Acoustic Token Model}. The reader is referred to \citet{GarciaSKP23} for full details of the model, and \Cref{apx:training} for a specification of hyperparameters used in training WhAM.

\paragraph{Acoustic Tokenizer.} The tokenizer takes as input an $\Nsec$-second audio snippet sampled at $\Nsamples$ Hz, and outputs a sequence of $\ell$ discrete tokens from a finite vocabulary $\Sigma$. A jointly-trained detokenizer will convert token sequences back into audio:
\begin{align*}
    T&\colon \R^{\Nsec \times \Nsamples} \to \Sigma^\ell \\
    T^{-1} &\colon \Sigma^{\ell} \to \R^{\Nsec \times \Nsamples}.
\end{align*}
VampNet uses a residual vector quantization approach known as the Descript Audio Codec (DAC, \citealt{KumarSLKK23}). At a high level, audio is tokenized in a temporal and hierarchical fashion, such that each interval of samples is replaced with a ``stack'' of tokens; this means that neighboring stacks of tokens correspond to contiguous intervals of samples in the audio. For example, the first five token stacks $(\sigma_1, \dots, \sigma_5)$ could correspond to the first 0.5 seconds of audio.

\paragraph{Masked Acoustic Token Model (MATM).} A bidirectional transformer $M$ is trained to perform the cloze task on acoustic token sequences. That is, each audio snippet in the pretraining dataset is tokenized, and then a bidirectional transformer is trained to predict a random subset of masked tokens.
\begin{equation*}
    M \colon (\Sigma \cup \{\MASK\})^{\ell} \to \Sigma^{\ell}
\end{equation*}

A pretrained MATM can be finetuned in various ways. Following \citet{GarciaSKP23}, we finetune using Low Rank Adaptation (LoRA, \citealt{HuSWALWWC22}).

\paragraph{Generation.} After training a tokenizer $T$, detokenizer $T^{-1}$ and a (possibly finetuned) MATM $M$, VampNet can be used to generate variations of given ``prompt'' audio snippets. This is done in the natural way, by randomly masking the tokenized audio; importantly, the masking scheme used in generation time does \emph{not} need to be uniformly random. For example, the scheme can leave (classically-detected) \emph{beats} unmasked, so as to preserve the rhythm of the prompt.
Rather than generating all masked tokens simultaneously (e.g. as in BERT, \citealt{BERT}), VampNet uses \emph{iterative parallel decoding} \citep{MaskGIT} wherein tokens are gradually ``unmasked'' in a sequence of forward passes through the model. 

\begin{figure}[ht]
\begin{center}
\centerline{\includegraphics[width=0.5\linewidth]{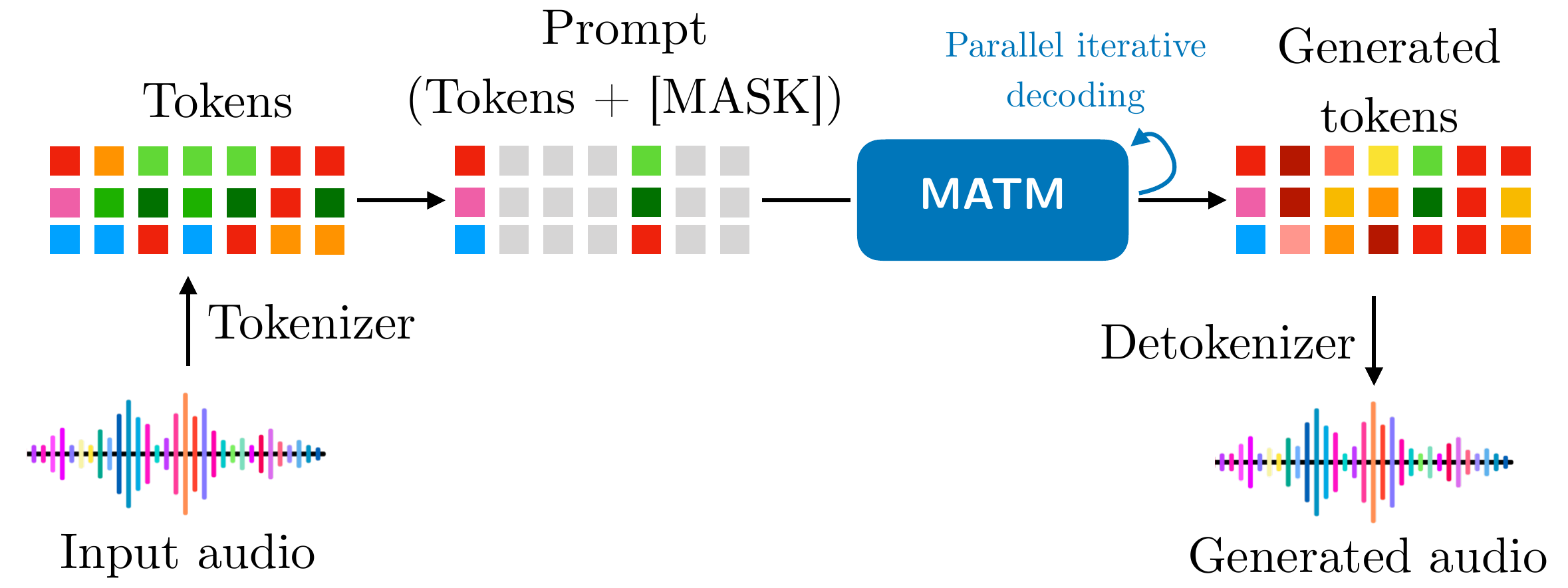}}
\caption{Overview of VampNet's generation pipeline. Input audio is first converted into a grid of tokens by the Tokenizer. These tokens are then partially masked to create a prompt. The Masked Acoustic Token Model (MATM) uses parallel iterative decoding to generate new tokens, which are finally converted back into audio by the Detokenizer. The colored squares represent acoustic tokens, with grey squares indicating masked positions.}
\label{fig:vampnet}
\end{center}
\vskip -0.2in
\end{figure}

\subsection{Data}\label{sec:data}
WhAM is trained by finetuning VampNet (\Cref{sec:VN}) on various datasets.
\begin{description}
	\item{FSD.} The Freesound Dataset \citep{FreeSound} consists of 50k human-labeled recordings. We used recordings with the \emph{animal} tag, which totaled 7h45m of audio.
	\item{AudioSet.} A dataset of two million human-labeled audio clips taken from YouTube \citep{AudioSet}. Of these, we used audio with the \emph{animal} tag, totaling at about 5 hours.
	\item{WMMS.} The Watkins Marine Mammal Sound Database \citep{WMMS} totaling 4h8m. It includes audio collected over seven decades in at least 67 sites around the world. Sperm whales are among the 51 species recorded.
    \item {BirdSet.}  An avian bioacoustics dataset curated for classification tasks \citep{Birdset}, totalling about 6,800 hours. Due to computational limits, we used a 110-hour subset of audio dense with vocalizations.
	\item{DSWP.} A dataset of 2507 annotated codas (1h26m) collected over thirteen years in a 2000km$^2$ area off the coast of Dominica. It consists of codas recorded using far-field boat-based hydrophones and noninvasive animal-borne tags.
	\item{CETI.} A more recent dataset of sperm whale vocalizations consisting of 7653 annotated codas (4h33m) collected similarly to DSWP.
\end{description}

The training of WhAM is split into two phases: (1) \emph{Domain adaptation}, in which the base VampNet is finetuned on FSD+AudioSet+WMMS for 500k iterations; (2) \emph{species-specific finetuning}, in which domain-adapted VampNet is finetuned on DSWP+CETI for an additional 500k iterations. Both phases follow the same (LoRA) finetuning procedure, but we find this split to be conceptually useful. 
Additional details are deferred to \Cref{apx:data}

\section{Results} \label{sec:results}
We evaluate WhAM through three complementary analyses. First, we assess the quality of WhAM's synthetic codas through quantitative metrics, specifically the Fr\'{e}chet Audio Distance (FAD, \citealt{FAD}) between generated and natural codas. Second, we conduct a perceptual study with expert marine biologists to evaluate how well our synthetic codas preserve the characteristic features of natural sperm whale vocalizations. Finally, we evaluate WhAM's learned representations on downstream classification tasks to investigate whether our model captures meaningful acoustic features of sperm whale communication.

\subsection{Fr\'{e}chet Distance of Audio Translation}\label{sec:fad}
\begin{figure*}[t]
    \centering
    \includegraphics[width=\linewidth]{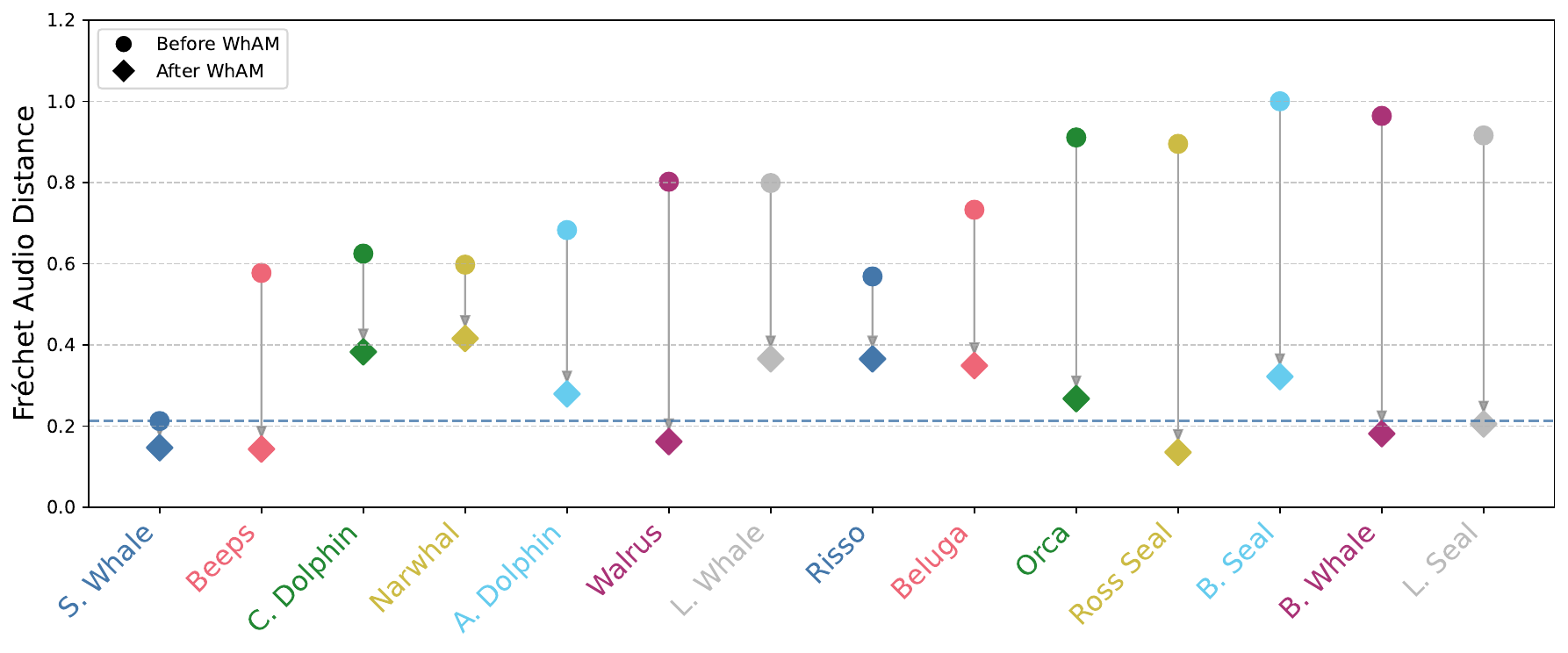}
    \caption{Normalized Fr\'{e}chet Audio Distance between sperm whale codas and various audio sources, before and after translation through WhAM. Lower FAD indicates greater acoustic similarity to natural codas. The horizontal line at 0.21 represents the baseline FAD between disjoint sets of natural codas. Full names of animals along with the number of samples from each can be found in \Cref{tab:marine_animals}.} 
    \label{fig:fad_results}
\end{figure*}
A key aspect of WhAM is its ability to ``translate'' audio inputs into the acoustic style of sperm whale codas. To evaluate this capability quantitatively, we measure the Fr\'{e}chet Audio Distance (FAD, \citealt{FAD}) between natural and WhAM-generated synthetic codas. FAD measures the similarity between two audio datasets by comparing embeddings of the audio signals; lower FAD indicates greater acoustic similarity between the datasets. 

FAD is computed using a given audio embedding model. We chose BirdNET \citep{BirdNet21} based on a principled calibration experiment that compared the sensitivity of five embedding models to the rhythmic patterns crucial to coda structure (\Cref{apx:fad_calibration}). We normalize FAD values by dividing by the maximum distance, scaling all values to $[0,1]$.
\Cref{fig:fad_results} portrays WhAM's translation ability using audio prompts from three domains:
\begin{enumerate}
    \item \emph{Natural codas (S. Whale)}: A disjoint set of codas produced by sperm whales. The FAD between disjoint sets of natural codas is 0.21 (rather than zero) due to variance in recording conditions, individual whales, and coda types. This establishes a baseline below which FAD fails to distinguish audio sources from natural variation: We therefore say that generated outputs are \emph{FAD-indistinguishable} when their FAD falls below 0.21. When passing natural codas through WhAM, we expect a slight decrease in FAD as WhAM regularizes inputs toward the mean embedding of its training distribution (a large dataset of diverse codas).
    
	\item \emph{Animal sounds}: Vocalizations from 12 species of marine mammals. \Cref{fig:fad_results} shows that WhAM consistently reduces the acoustic distance to natural codas, effectively translating these diverse inputs into the acoustic style of sperm whale codas. WhAM-generated outputs of four species are FAD-indistinguishable from natural codas.

    \item \emph{Digital ``beeps''}: Artificial audio generated by initializing an array of zeros and randomly selecting points to assign a peak amplitude of 1. Remarkably, beeps and natural codas have approximately the same post-WhAM FAD. This may be because beeps' sparse structure (mostly silence with isolated peaks) gives WhAM freedom to infill patterns close to the mean embedding of codas, while natural codas with minimal silence constrain the model's regularization but start closer to the target distribution.

\end{enumerate}

The results demonstrate WhAM's remarkable translation capability: five diverse audio sources (four non-whale species and digital beeps) become FAD-indistinguishable from natural sperm whale codas after processing. This success across varied inputs suggests that WhAM has learned a robust representation of the essential acoustic properties that define sperm whale communication.

\subsection{Expert Perceptual Study}\label{sec:listening}
\begin{figure*}[t]
\begin{minipage}[t]{0.48\textwidth}
    \centering
    \includegraphics[width=\textwidth]{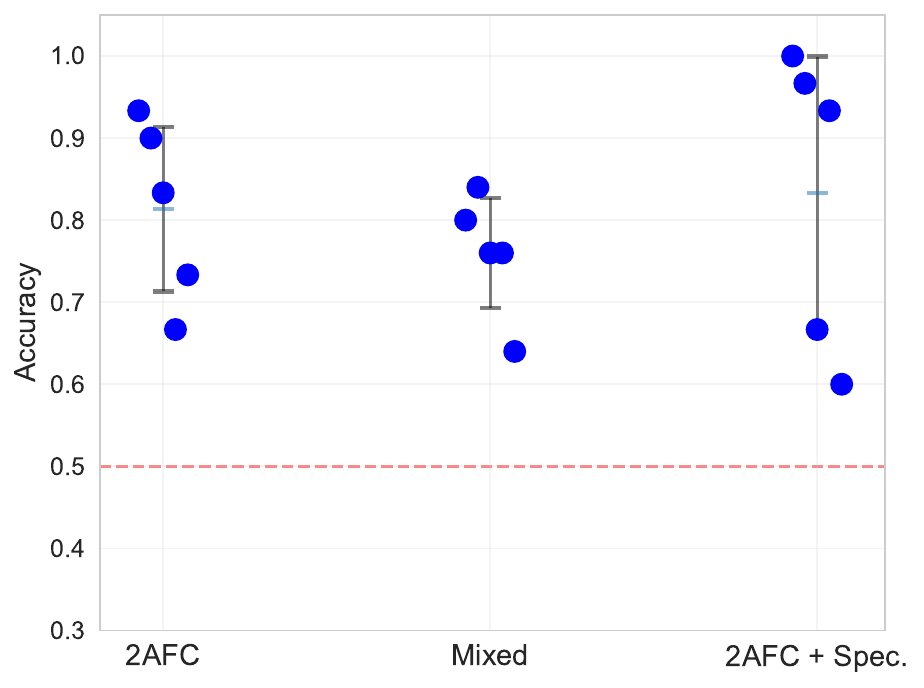}
    \caption{Expert performance on audio-only 2AFC (Task 1), mixed classification (Task 2), and spectrogram-assisted 2AFC (Task 3). Error bars show standard deviation across experts. While all tasks elicited above-chance performance (dashed line), spectrogram analysis showed the greatest variability between experts ($\sigma=0.17$). Task 1 and 3 had 30 pairs each, Task 2 had a collection of 25 samples.}
    \label{fig:tasks}
\end{minipage}
\hfill
\begin{minipage}[t]{0.48\textwidth}
    \centering
    \includegraphics[width=\textwidth]{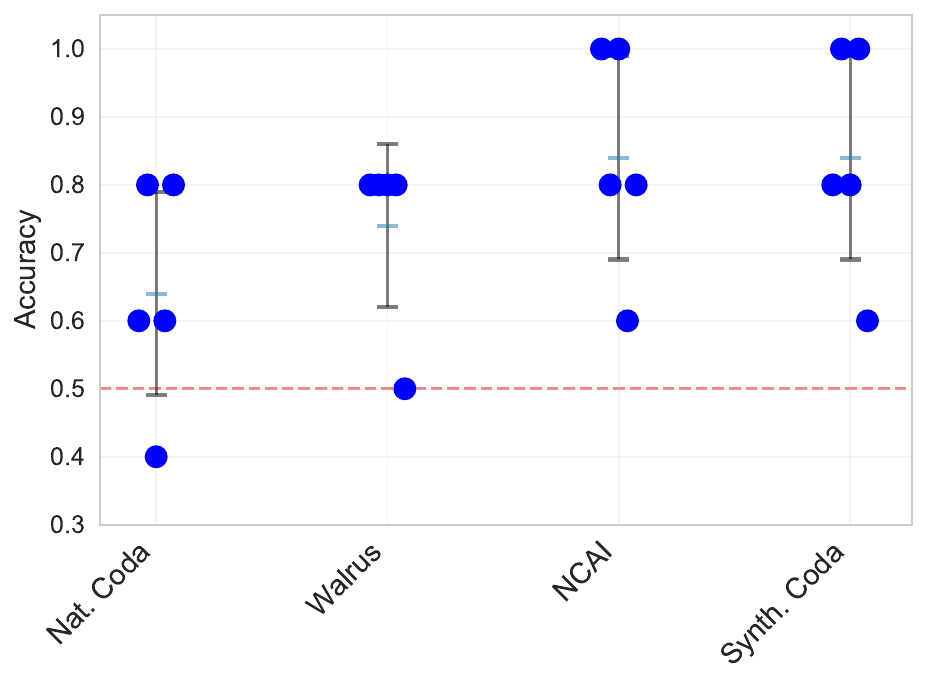}
    \caption{Accuracy in mixed classification (Task 2) for different input domains. Natural codas (left) were misclassified as synthetic 36\% of the time. The remaining columns depict performance on synthetic codas generated by WhAM from walrus vocalizations, non-coda acoustic impulses, and codas (respectively). There were five synthetic codas from each domain, plus ten natural codas for a total of 25 items.}
    \label{fig:domains}
\end{minipage}
\end{figure*}
To evaluate the perceptual quality of WhAM's synthetic codas, we conducted a comprehensive study with domain experts to assess how well our generated outputs match natural sperm whale vocalizations with respect to a human-expert distinguisher. This study measures both audio-only and spectrogram-based discrimination, while also gathering qualitative insights about specific acoustic features that distinguish synthetic from natural codas. Additional details are deferred to \Cref{apx:listening}.

\paragraph{Expert backgrounds.} Five academic experts participated in the perceptual study. Three identified as marine biologists, and two as underwater acoustics specialists. They listed between 3 and 20 years of experience working with coda audio (field recordings), manual detection and classification, and the development of automatic detection systems. All experts had experience analyzing coda audio and spectrograms, which are the two media through which the experiment was carried out.

\paragraph{Experiment design.} We designed a four-task study to be completed sequentially by each expert:
\begin{enumerate}
	\item \textbf{Audio-only two-alternative forced choice (2AFC):} Experts compared pairs of codas (one natural, one synthetic) in audio-only conditions, and were asked to identify the synthetic coda. Synthetic codas were generated by WhAM using the paired natural coda as input.
\item \textbf{Mixed Collection Classification:} Experts classified clips as natural or synthetic. Clips were either natural codas, or synthetic codas generated from different sources: natural codas, digital beeps,\footnote{i.e., an artificial sequence of clicks} or walrus vocalizations \citep{WMMS}. False positives (natural misclassified) and false negatives (synthetic undetected) were measured.
	\item \textbf{Spectrogram-assisted 2AFC:} Experts repeated the first task while visualizing audio with software of their choice. The experts were given the exact same samples as in the first task, ensuring direct comparability between audio-only and spectrogram-aided performance. This task mirrored real-world analysis workflows while quantifying the perceptual ``advantage'' of multimodal inspection.
	\item \textbf{Qualitative assessment:} Experts were given five representative samples of synthetic codas. They were then asked questions about how well synthetic codas captured / missed characteristics of natural codas, whether any non-natural patterns appeared in synthetic codas, and which features did they use to distinguish between codas in each of the previous tasks.
\end{enumerate}

Fleiss's $\kappa$ quantified inter-expert agreement \citep{Fleiss71}, and accuracy was calculated relative to ground-truth labels. Task order was chosen towards minimizing bias (audio-first to avoid visual priming), with background information collected in a final section. The samples used in the experiment are attached to this submission as supplementary material. Experimental details are in Appendix \ref{apx:listening}.

\paragraph{Quantitative analysis.}
Experts achieved 81\% accuracy ($\kappa=0.41$), in audio-only 2AFC (Task 1), rising marginally to 83\% ($\kappa=0.41$) with spectrograms visualized (Task 3). This 2\% improvement suggests WhAM's synthetic codas lack glaring spectro-temporal artifacts detectable by trained analysts. As expected, accuracy with spectrograms was generally better per-expert, with one expert's performance dramatically increasing from 66\% to 93\% (another expert even achieved a perfect score). Surprisingly, one expert's performance \emph{decreased} from 83\% in Task 1 to 66\% in Task 3; comments in the qualitative section did not suggest an explanation.

Performance varied substantially across tasks and among experts (\Cref{fig:tasks}). The most experienced expert ranked highest in both 2AFC tasks, but not in mixed classification. These variations reflect diverging expert strategies---some focused on inter-click patterns, others on spectral properties: ``rhythm'' to quote one expert, versus ``DC offsets'' and ``inter-pulse structures'' \citep{MohlWMHL03} to quote others.

Misclassification rates in Task 2 (\Cref{fig:domains}) revealed WhAM's efficacy in acoustic translation: on average, experts correctly flagged walrus-to-coda audio only 75\% of the time---less than digital beeps or coda-to-coda outputs of WhAM. For one expert, walrus-to-coda audio was detected only 50\% of the time (random chance). 

Fleiss's $\kappa$ values (0.41--0.44) indicated moderate agreement across tasks, with experts showing greatest consensus on mixed classification ($\kappa=0.44$). Performance on spectrogram-aided 2AFC performance was the most diverse---one expert achieved perfect performance while another approached chance (60\%).

\paragraph{Qualitative feedback.}
Synthetic codas successfully replicated key acoustic features of natural codas. Most experts noted preservation of \emph{rhythm}, referred to as inter-click intervals (ICI); that is, clicks occur ``at the right time'' in synthetic codas. Additionally, one expert answered that ``spectral components'' were overall preserved in synthetic codas.

That said, experts identified missing components which can be partitioned into three categories:
\begin{itemize}
	\item \emph{Within a single click:} Some clicks ``came on and disappeared too strongly,'' had ``varying amplitude [within a single coda],'' and ``inverted peaks.''
    On a spectral level, an expert answered that clicks were too ``broadband'' compared to natural clicks which have a low-frequency bias.
	\item \emph{Rhythmic/temporal:} One expert noted that the timing of clicks fit echolocation moreso than codas. (See \Cref{apx:codas} for how they differ.)
	\item \emph{Recording-level anomalies:} One expert noted a ``DC offset'' which they described as the unrealistic background noise on synthetic codas. Similarly, another noted that background noise in synthetic codas oscillated too much.
\end{itemize}

Based on this assessment, we present in \Cref{apx:listener-guide} a guide to the similarities and differences between natural and synthetic codas, supplemented by annotated spectrograms.

\subsection{Utility of embeddings for downstream tasks}\label{sec:downstream}
We test whether WhAM's internal representations capture meaningful features of sperm whale vocalizations through three downstream classification tasks. For each task, we train a small (two-layer) classifier head that takes coda embeddings as input. We compare WhAM to naive random-embedding and majority-class baselines, as well as AVES \citep{Hagiwara23}, a self-supervised model achieving state-of-the-art performance on bioacoustic classification tasks. Full details of the experimental setup are deferred to \Cref{apx:downstream}.

The downstream tasks are:
\begin{enumerate}
    \item \emph{Coda detection}: Given a snippet of audio, determine whether it contains a coda. The classifier is trained on positive (coda) and negative (no coda) snippets, with negative examples drawn from the same recording conditions to ensure the model learns coda features rather than recording artifacts. %
    
    \item \emph{Rhythm type}: Given a snippet of audio, classify its temporal pattern. Rhythm of inter-click intervals serves as a key axis for classification of sperm whale codas in cetacean research \cite{ShulzWGR11,BermantBWGG19,SharmaGPGRTA24}.
    
    \item \emph{Social unit classification}: The lowest level of sperm whale social structure are called social units (SU) and have stable, matrilineally-related membership of females and their young \citep{ChristalWL98}. Importantly, all SUs in DSWP+CETI belong to the same vocal clan and thus share a common repertoire of coda types, making this more of a speaker identification task than dialect classification.\footnote{By analogy to human language, consider the task of classifying speakers by city of origin. It would be easier to distinguish between speakers from cities that use different dialects (simply classify the dialect). Importantly, in our data, all speakers use the same dialect.}

    \item \emph{Vowel type}: Given a coda recording, classify its vowel-like pattern \citep{Vowels}.
\end{enumerate}

\Cref{tab:downstream} shows classification accuracies for each task. As expected, non-generative embeddings specifically designed for acoustic classification tasks outperform outperform WhAM. We view those as a ceiling ``sanity check'' than a baseline. AVES and BirdNET perform particularly well on more specialized bio-acoustic tasks, due to the fact that they were both trained on large amounts of animal vocalizations. Notably, WhAM's representations are useful despite being trained only for generation, outperforming both naive baselines. This suggests that meaningful acoustic features emerge naturally from training for coda generation, even without explicit supervision for these tasks.

We conducted an ablation study to assess how fine-tuning affects embedding quality by evaluating different WhAM variants with specific components removed (detailed in \Cref{apx:downstreamAblation}). The results reveal that fine-tuning did not significantly alter WhAM's downstream utility compared to base VampNet embeddings, despite WhAM's specialization on whale codas. However, as shown in \Cref{apx:fad_ablation}, species-specific fine-tuning was essential for enabling WhAM's core capability of translating audio into sperm whale vocalization acoustics.

\begin{table}[t]
\caption{Classification accuracies (\%) of different audio embeddings. Each classifier head was trained using three different random seeds, with mean$\pm$stderr reported. The Random baseline uses a randomly initialized AVES model (training only the classifier), while Majority predicts the most common class.}
\label{tab:downstream}
\vskip 0.15in
\begin{center}
\begin{small}
\begin{sc}
\begin{tabular}{lcccccc}
\toprule
Task & WhAM & \multicolumn{2}{c}{Baseline} & AVES & BirdNET & CLAP \\
\cmidrule(lr){3-4}
& & Rand. & Maj. & & & \\
\midrule
Detection & 91.3$\pm$0.2 & 60.9 & 60.9 & 92.8$\pm$0.1 & 93.0$\pm$1.0 & \textbf{96.8}$\pm$1.4 \\
Rhythm & 87.4$\pm$1.6 & 66.3 & 60.9 & 90.4$\pm$1.6 & 88.6$\pm$0.2 & \textbf{92.4}$\pm$2.4 \\
Social Unit & 70.5$\pm$5.6 & 42.5 & 35.1 & 92.0$\pm$0.7 & \textbf{93.2}$\pm$0.1 & 85.5$\pm$1.4 \\
Vowel & 85.2$\pm$2.5 & 66.3 & 66.3 & \textbf{91.8}$\pm$2.9 & 85.9$\pm$4.6 & 84.3$\pm$0.9 \\
\bottomrule
\end{tabular}
\end{sc}
\end{small}
\end{center}
\vskip -0.1in
\end{table}

\section{Limitations and Future Work}\label{sec:future}
The most immediate technical limitation concerns the audio codec architecture. Our current implementation only finetunes the MATM while keeping the codec fixed (see \Cref{sec:VN}). This design choice, while computationally efficient, may limit the model's ability to capture nuanced acoustic features specific to sperm whale vocalizations. For instance, the recently discovered vowel-like patterns in the 3.7--5.7kHz band \citep{Vowels} may be inadequately represented by a codec primarily trained on human music. Future work could explore either finetuning the entire codec or developing specialized codecs for bioacoustic signals.

Expert feedback (\Cref{sec:listening}) highlighted specific limitations in click generation: unnatural onset and decay patterns, inconsistent background noise, and click properties more reminiscent of echolocation than communication codas. These limitations might be addressed through architectural modifications, such as incorporating adversarial components \citep{BegusLG23} or introducing specialized modules that leverage domain knowledge about sperm whale click structure. Notably, the observation about echolocation-like properties led to an unexpected finding in our dataset preparation: the presence of echolocation sequences in datasets intended for communication codas. This discovery highlights a broader challenge in bioacoustics research—the difficulty of building clean, well-labeled datasets at scale. Future work should focus on developing robust methods for distinguishing between different types of vocalizations, perhaps by leveraging existing detection systems \citep{BermantBWGG19}.

These data quality challenges underscore the importance of thorough evaluation protocols. Expanding the expert panel would provide more robust perceptual assessments, though we acknowledge the practical challenges in recruiting specialists in sperm whale vocalizations. Additionally, developing more principled evaluation methods---and meta-evaluating these---would help establish standardized benchmarks for bioacoustic generation tasks.

While our results demonstrate impressive performance with relatively small datasets---orders of magnitude smaller than typical in modern acoustic model training---scaling up the training data could yield substantial improvements. This would require significant effort in aggregating and preprocessing additional sperm whale datasets, as our experience with DSWP+CETI highlighted the technical challenges involved in preparing bioacoustic data for machine learning pipelines.

Looking beyond technical improvements, future work could explore unsupervised learning approaches to uncover new coda features, following the success of similar approaches in bioacoustics \citep{Vowels}. This could lead to discoveries about sperm whale communication that complement traditional analytical methods while providing new directions for improving generative models of animal vocalizations.

Our methodological framework---from the two-phase training approach to the expert evaluation protocol---could be adapted for studying other animal communication systems. Our experience suggests that success will require careful attention to species-specific acoustic features and close collaboration with domain experts who can identify subtle but important characteristics of vocalizations.

The gap between generating vocalizations and understanding their meaning remains vast. While WhAM represents the first attempt at acoustic translation in the context of sperm whale communication, future work should explore ways to bridge this semantic gap while maintaining minimal assumptions about the underlying communication system.

\begin{ack}
    We thank Guy Gubnitsky, Yaly Mevorach, and Pernille Tonnesen for volunteering their time and expertise to the expert perceptual study. We thank anonymous reviewers for helpful comments, and in particular reviewer fPAV for suggesting adding BirdSet \citep{Birdset} to the domain adaptation stage.

    This study was funded by Project CETI via grants from Dalio Philanthropies and Ocean X; Sea Grape Foundation; Virgin Unite and Rosamund Zander/Hansjorg Wyss through The Audacious Project: a collaborative funding initiative housed at TED. This work was supported, in part, by National Science Foundation Award \# 2222369.

    Fieldwork for The Dominica Sperm Whale Project, which produced many of the coda recordings used in this work, was supported by a FNU fellowship for the Danish Council for Independent Research supplemented by a Sapere Aude Research Talent Award (1325-00047A), a Carlsberg Foundation expedition grant (CF14-0789), a grant from Focused on Nature, two Explorer Grants from the National Geographic Society (WW-218R-17 and NGS-64863R-19), and supplementary grants from the Arizona Center for Nature Conservation, Quarters For Conservation to SG, with  supplemental funding from the Dansk Akustisks Selskab, Oticon Foundation, and the Dansk Tennis Fond to Pernille Tonnesen. Further funding was provided by Discovery and Equipment grants from the Natural Sciences and Engineering Research Council of Canada (NSERC) to Hal Whitehead of Dalhousie University and a FNU large frame grant and a Villum Foundation Grant (13273) to Peter Madsen of the Marine Bioacoustics Lab at Aarhus University. We thank the Chief Fisheries Officers and the Dominica Fisheries Division officers for research permits and their collaboration in data collection; all the crews of R/V Balaena and The DSWP team for data collection, curation, and annotation of audio recordings and photoID; as well as Dive Dominica, Al Dive, W.E.T. Dominica, and Wacky Rollers for logistical support while in Dominica. We are grateful to Kristian Beedholm of the Marine Bioacoustics Lab at Aarhus University for CodaSorter; as well as Mark Johsson and Peter Tyack for in-kind contributions of Dtags and associated code.
\end{ack}

\bibliography{icml2025}
\bibliographystyle{plainnat}

\appendix
\section*{NeurIPS Paper Checklist}

\begin{enumerate}

\item {\bf Claims}
    \item[] Question: Do the main claims made in the abstract and introduction accurately reflect the paper's contributions and scope?
    \item[] Answer: \answerYes{} %
    \item[] Justification: Fr\'{e}chet Audio Distance evaluations can be found in \Cref{sec:fad}. Perceptual studies with expert marine biologists can be found in \Cref{sec:listening}. Experiments on downstream classification are reported in \Cref{sec:downstream}.

\item {\bf Limitations}
    \item[] Question: Does the paper discuss the limitations of the work performed by the authors?
    \item[] Answer: \answerYes{} %
    \item[] Justification: Yes, in \Cref{sec:future}.

\item {\bf Theory assumptions and proofs}
    \item[] Question: For each theoretical result, does the paper provide the full set of assumptions and a complete (and correct) proof?
    \item[] Answer: \answerNA{} %
    \item[] Justification: N/A.

    \item {\bf Experimental result reproducibility}
    \item[] Question: Does the paper fully disclose all the information needed to reproduce the main experimental results of the paper to the extent that it affects the main claims and/or conclusions of the paper (regardless of whether the code and data are provided or not)?
    \item[] Answer: \answerYes{} %
    \item[] Justification: Details on model training are in \Cref{apx:methods}. Details on the experimental setup are in \Cref{apx:experiments}.

\item {\bf Open access to data and code}
    \item[] Question: Does the paper provide open access to the data and code, with sufficient instructions to faithfully reproduce the main experimental results, as described in supplemental material?
    \item[] Answer: \answerNo{} %
    \item[] Justification: We will release the model weights, and the code used for training and evaluating the model upon publication. We will attempt to release as much of the data as possible, however we note that the data was collected by a large collaboration of marine biologists over several decades, and so we cannot commit to getting their consent to publish by publication time.

\item {\bf Experimental setting/details}
    \item[] Question: Does the paper specify all the training and test details (e.g., data splits, hyperparameters, how they were chosen, type of optimizer, etc.) necessary to understand the results?
    \item[] Answer: \answerYes{} %
    \item[] Justification: Yes, in \Cref{apx:methods,apx:experiments}.

\item {\bf Experiment statistical significance}
    \item[] Question: Does the paper report error bars suitably and correctly defined or other appropriate information about the statistical significance of the experiments?
    \item[] Answer: \answerYes{} %
    \item[] Justification: All figures depict error bars or standard error figures, with the exception of \Cref{fig:fad_results,fig:generative_ablations,tab:fad_calibration}. For \Cref{fig:fad_results,fig:generative_ablations} repeating the experiments with multiple seeds would have been computationally prohibitive due to the amount of categories in each experiment. For \Cref{tab:fad_calibration}, this is a minor, supplementary used to justify an experimental design choice, rather than a main component of the paper.

\item {\bf Experiments compute resources}
    \item[] Question: For each experiment, does the paper provide sufficient information on the computer resources (type of compute workers, memory, time of execution) needed to reproduce the experiments?
    \item[] Answer: \answerYes{} %
    \item[] Justification: The computer resources needed to train and finetune the model are reported in \Cref{apx:training}. The resources needed to conduct the experiments (after training the model) are reported in \Cref{apx:experiments}.
    
\item {\bf Code of ethics}
    \item[] Question: Does the research conducted in the paper conform, in every respect, with the NeurIPS Code of Ethics \url{https://neurips.cc/public/EthicsGuidelines}?
    \item[] Answer: \answerYes{} %
    \item[] Justification: We adhere to the NeurIPS Code of Ethics.

\item {\bf Broader impacts}
    \item[] Question: Does the paper discuss both potential positive societal impacts and negative societal impacts of the work performed?
    \item[] Answer: \answerYes{} %
    \item[] Justification: In \Cref{sec:impact}
    
\item {\bf Safeguards}
    \item[] Question: Does the paper describe safeguards that have been put in place for responsible release of data or models that have a high risk for misuse (e.g., pretrained language models, image generators, or scraped datasets)?
    \item[] Answer: \answerNA{} %
    \item[] Justification: N/A.

\item {\bf Licenses for existing assets}
    \item[] Question: Are the creators or original owners of assets (e.g., code, data, models), used in the paper, properly credited and are the license and terms of use explicitly mentioned and properly respected?
    \item[] Answer: \answerYes{} %
    \item[] Justification: We adhere to the license of all external code we use. The data used was either collected by us (the authors), or is used in accordance to the license of the dataset.

\item {\bf New assets}
    \item[] Question: Are new assets introduced in the paper well documented and is the documentation provided alongside the assets?
    \item[] Answer: \answerYes{} %
    \item[] Justification: We provide as much data as appropriate and possible of the new datasets we used, without de-anonymizing ourselves. The training method and model are properly documented in \Cref{apx:methods}.

\item {\bf Crowdsourcing and research with human subjects}
    \item[] Question: For crowdsourcing experiments and research with human subjects, does the paper include the full text of instructions given to participants and screenshots, if applicable, as well as details about compensation (if any)? 
    \item[] Answer: \answerYes{} %
    \item[] Justification: In \Cref{apx:listening}.

\item {\bf Institutional review board (IRB) approvals or equivalent for research with human subjects}
    \item[] Question: Does the paper describe potential risks incurred by study participants, whether such risks were disclosed to the subjects, and whether Institutional Review Board (IRB) approvals (or an equivalent approval/review based on the requirements of your country or institution) were obtained?
    \item[] Answer: \answerNo{} %
    \item[] Justification: The perceptual listening study presented no risk to participants. Participants were asked to evaluate audio similar to those they regularly analyze in their professional work. No personal data was collected beyond coarse background questions and the assessment itself (see \Cref{apx:experiments}), and no compensation was provided. The task falls within the participants' scope of their normal research activities and expertise.

\item {\bf Declaration of LLM usage}
    \item[] Question: Does the paper describe the usage of LLMs if it is an important, original, or non-standard component of the core methods in this research? Note that if the LLM is used only for writing, editing, or formatting purposes and does not impact the core methodology, scientific rigorousness, or originality of the research, declaration is not required.
    \item[] Answer: \answerNA{} %
    \item[] Justification: N/A.

\end{enumerate}

\section{Broader impacts}\label{sec:impact}
Our work on modeling sperm whale communication has potential implications for both scientific understanding and conservation efforts. Historically, advances in understanding cetacean communication have played crucial roles in conservation---notably, the discovery of humpback whale song by \citet{PayneM71} contributed significantly to public awareness and the subsequent ``Save the Whales'' movement \citep{Feldman21,Campagna22,Comuzzo23}. While we maintain that sperm whales deserve protection regardless of our ability to understand their communication, we recognize that deeper scientific understanding often catalyzes public engagement with conservation efforts.

Our model's capabilities might naturally suggest applications in behavioral experiments through playback studies. This is particularly tempting given that sperm whales often produce codas simultaneously---a behavior that our bidirectional model could theoretically capture by conditioning on one whale's clicks while generating the overlapping clicks of another. However, we strongly caution against such applications at this stage. Without a deeper understanding of coda semantics and functionality, playback experiments using synthetic vocalizations could have unintended and potentially harmful consequences for these social marine mammals. Instead, we propose that this work demonstrates the potential of learning from passive acoustic observation—studying these remarkable animals through careful listening rather than active intervention. With this approach, this work could potentially play a role in assisting efforts to reinforce existing protections or create new legal protections for whales \citep{ceti-law}.

As noted in \Cref{sec:future}, our methodological framework could extend beyond sperm whales, potentially benefiting research on other marine mammals and, more broadly, any species that communicates acoustically. This scalability is particularly relevant as biodiversity monitoring becomes increasingly critical in the face of environmental changes. However, our experience underscores the importance of deep collaboration with domain experts---the success of this work relied on guidance from marine biologists and acousticians with decades of experience studying sperm whales. We  encourage future work in this direction to similarly prioritize partnerships with species-specific domain experts, as their insights are crucial for both model development and responsible deployment.

\section{Preliminaries on sperm whale codas}\label{apx:codas}
\begin{figure}[thb]
    \centering
    \includegraphics[width=\linewidth]{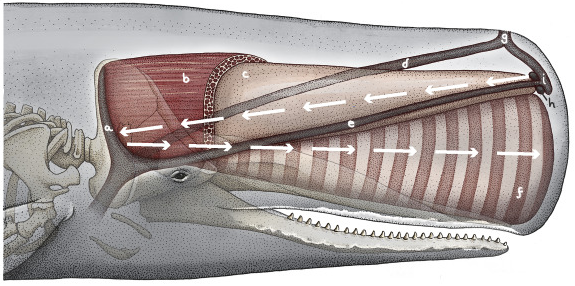}
    \caption{
        The sperm whale head contains the spermaceti organ (c), a cavity filled with almost 2kL of wax-like liquid, and the junk compartment (f), comprising a series of wafer-like bodies believed to act as acoustic lenses. The spermaceti organ and junk act as two connected tubes, forming a bent, conical horn of about 10m in length and 0.8m aperture in large mature males. The sound emitted by the phonic lips (i) in the front of the head is focused by traveling through the bent horn, producing a flat wavefront at the exit surface. Reproduced with permission (\citealt{AndreasEtAl22}, \textcopyright{} Alex Boersma 2021).
    }
    \label{fig:codas1}
\end{figure}
\begin{figure}[hbt]
    \centering
    \begin{minipage}[t]{0.65\linewidth}
        \centering
        \includegraphics[width=\linewidth]{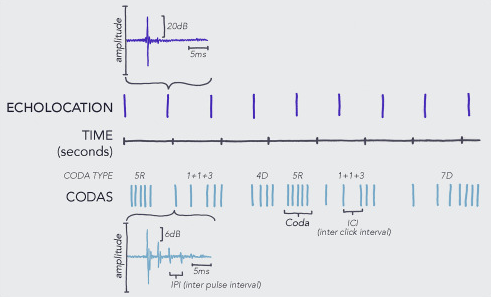}
    \end{minipage}
    \hfill
    \begin{minipage}[t]{0.32\linewidth}
        \centering
        \includegraphics[width=\linewidth]{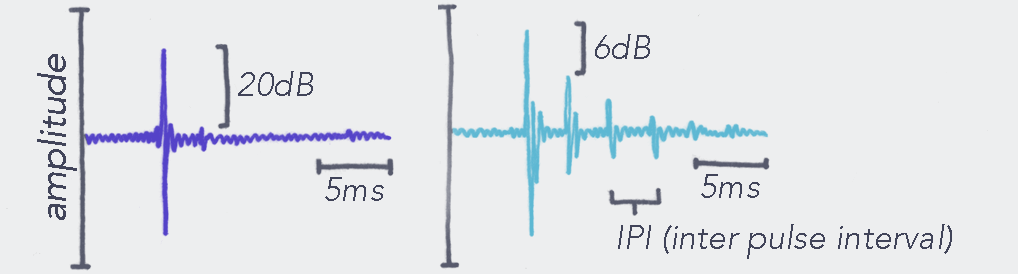}
    \end{minipage}
    \caption{
        \textbf{Left:} Typical temporal structure of sperm whale echolocation and coda clicks. Echolocation signals are produced with consistent inter-click intervals (of approximately 0.4s) while coda clicks are arranged in stereotypical sequences called ``codas'' lasting less than 2s. Codas are characterized by the different number of constituent clicks and the intervals between them (called inter-click intervals). Codas are typically produced in multi-party exchanges that can last from about 10s to over half an hour. Each click, in turn, presents itself as a sequence of equally spaced pulses, with inter-pulse interval of an order of 3--4ms in an adult female, which is the result of the sound reflecting within the spermaceti organ.
        \textbf{Right:} Typical structure of echolocation (dark blue, left) and coda clicks (light blue, right). When observed as a waveform zoomed into a single click the type types of clicks differ observably in structure. There is far greater attenuation between the first and second pulse of an echolocation click, then the coda clicks. Further, the coda clicks resonate more in the nose of the sperm whale creating additional pulses after the first one for coda clicks. Reproduced with permission (\citealt{AndreasEtAl22}, \textcopyright{} Alex Boersma 2021).
    }
    \label{fig:codas2}
\end{figure}
Sperm whales have evolved remarkable acoustic capabilities. \Cref{fig:codas1} illustrates the key anatomical and acoustic aspects of these capabilities, which form the basis for their complex  communication system.

Sperm whales live in a multileveled social structure with female lines living together in 'units' with stable membership \citep{Whitehead03}. Early acoustic research proposed that codas might serve as individual signatures \citep{WatkinsS77}, but subsequent studies instead suggested that different coda types may have distinct functions \citep{AntunesSGWGR11}, and that variation of coda usage among units suggested a function in unit-level social identity \citep{MooreWT93,WeilgartW93,WeilgartW97}. Even when living in the same waters, whales from different social units will only associate with units which share a similar repertoire of codas. This social segregation based on acoustic similarity was used to delineate the highest level of social organization which structures their populations, the vocal clan; and that codas function as symbolic markers of these cultural groups \citep{RendellW03,GeroBWM16,HershEtAl22}. Importantly, there is good evidence that these distinct dialects of codas, with variation in number of clicks, as well as rhythm and tempo, are the product of social learning, and not genetically inherited \citep{CantorW15,RendellMDBW12}.

The clicks produced by sperm whale can be generally classified as either \emph{ecolocation} or \emph{coda} clicks.
Echolocation clicks which function in navigation and hunting in the dark, wherein echoes of the clicks return and are interpreted by the whales in the darkness of the deep ocean, much like bats in the night sky. Conversely, coda clicks are thought to function in communication between whales and are exchanged between whales or groups of whales in social contexts at the onset of dives, during shallow dives near the surface, and during large social interactions.

Echolocation signals are produced with consistent inter-click intervals while coda clicks are arranged in stereotyped, rhythmic sequences called ``codas'' lasting less than 2 seconds. Codas are characterized by the different number of constituent clicks and the intervals between them (called inter-click intervals or ICIs). Rhythmic patterns and tempo of clicks define coda ‘types’, which are often given descriptive names. For example, a 1+1+3 coda is click-pause-click-pause-click-click-click (\Cref{fig:codas2}).

\section{A listener's guide to codas}\label{apx:listener-guide}
Building on findings from our Expert Perceptual Study (\Cref{sec:listening}), we present a short guide detailing perceivable similarities and differences between natural and synthetic codas. We note that, unlike the Expert Study, this guide was developed by the authors under no time constraints, and with unrestricted aid of spectrograms and familiarity with model internals. This Listener's Guide to Codas is structured as a unifying Theme, followed by four Variations each isolating a specific cue.\footnote{With apologies to \cite{Britten46}.}
For a broad-audience listener's guide to whale (albeit humpback) vocalizations, see \cite{PayneSongs}.

\paragraph{Theme.}
Synthetic codas generated by WhAM can be evaluated both visually and acoustically, using the same structural cues that characterize authentic sperm whale clicks. Each natural coda click typically consists of a sequence of equally spaced pulses, with an inter-pulse interval (IPI) of approximately 3--4~ms in adult females. This is a consequence of internal reflections within the spermaceti organ.

\paragraph{Variation A: Balance.}
DC offset (a shift of the waveform away from being centered at zero) does sometimes occur when recording sperm whales in the wild, particularly when using handheld recording systems which run off batteries and a constant DC voltage. It is often consistent, while synthetically generated clips will have quite a ``wavy'' offset. It is however interesting to note that WhAM picked up on this feature of the authentic waveforms. In addition, sperm whales do not vary the amplitude dramatically between sequential clicks within codas, while WhAM generated codas sometimes do.

\paragraph{Variation B: Frequency.}
Using a spectrogram, one can see that the frequency content of synthetic clicks is more uniform. In \Cref{fig:shane-fig4}, one can compare the shape of spectrograms for otherwise relatively similar clicks and note that the shape is more uniform and consistent both in time and frequency for synthetic clicks (bottom as strong orange rectangles) compared to authentic clicks which trail off both as frequency increases and across time (top, more pointed at top, with far less yellow above 10kHz, and rough along the right side). In addition, you can also observe the variation in amplitude across synthetic clicks in the waveform, but a consistent amplitude in the waveform of the authentic clicks (as described above).

\paragraph{Variation C: Structure.}
Authentic sperm whale clicks, especially coda clicks, contain the typical multiplulsed structure with a detectable inter-pulse-interval created by the head of the sperm whale and the path of the sound as it is generated  (outlined above). Synthetic clicks often did not have a realistic structure either by having no pulsed structure (center of \Cref{fig:shane-fig5}) or an exaggerated one (right of \Cref{fig:shane-fig5}). While some of these effects occur in authentic clicks based on the angle of recording relative to the axis of the body of the whale making the sounds, the synthetic clicks rarely had realistic structure within clicks.

\paragraph{Variation D: Listening alone.}
Taken together, these waveform- and spectrogram-based cues are sometimes audible \emph{even without visual aids}. A trained ear could identify synthetic codas based on subtle irregularities in amplitude, spectral consistency, and the absence of realistic multipulsed structure.

\begin{figure*}[bht]
    \centering
    \includegraphics[width=0.9\linewidth]{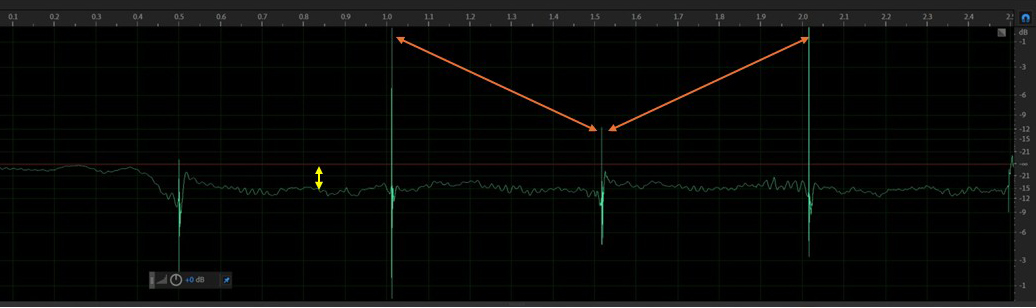}
    \caption{Sample of a synthetic coda generated by WHAM with the variable DC offset dissimilar to natural recordings (yellow arrow) and the dramatic variation in amplitude between sequential clicks (orange arrows).} 
    \label{fig:shane-fig3}
\end{figure*}
\begin{figure*}[bht]
    \centering
    \includegraphics[width=\linewidth]{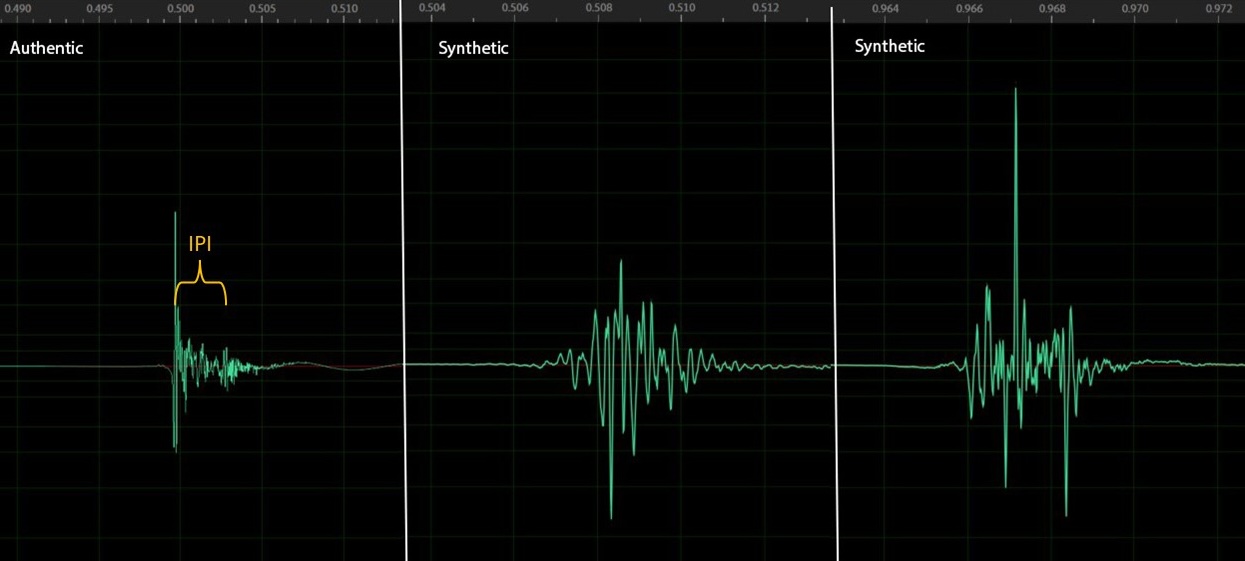}
    \caption{Pulse structure of authentic (real) and synthetic clicks.}
    \label{fig:shane-fig5}
\end{figure*}
\begin{figure*}[hbt]
    \centering
    \includegraphics[width=0.9\linewidth]{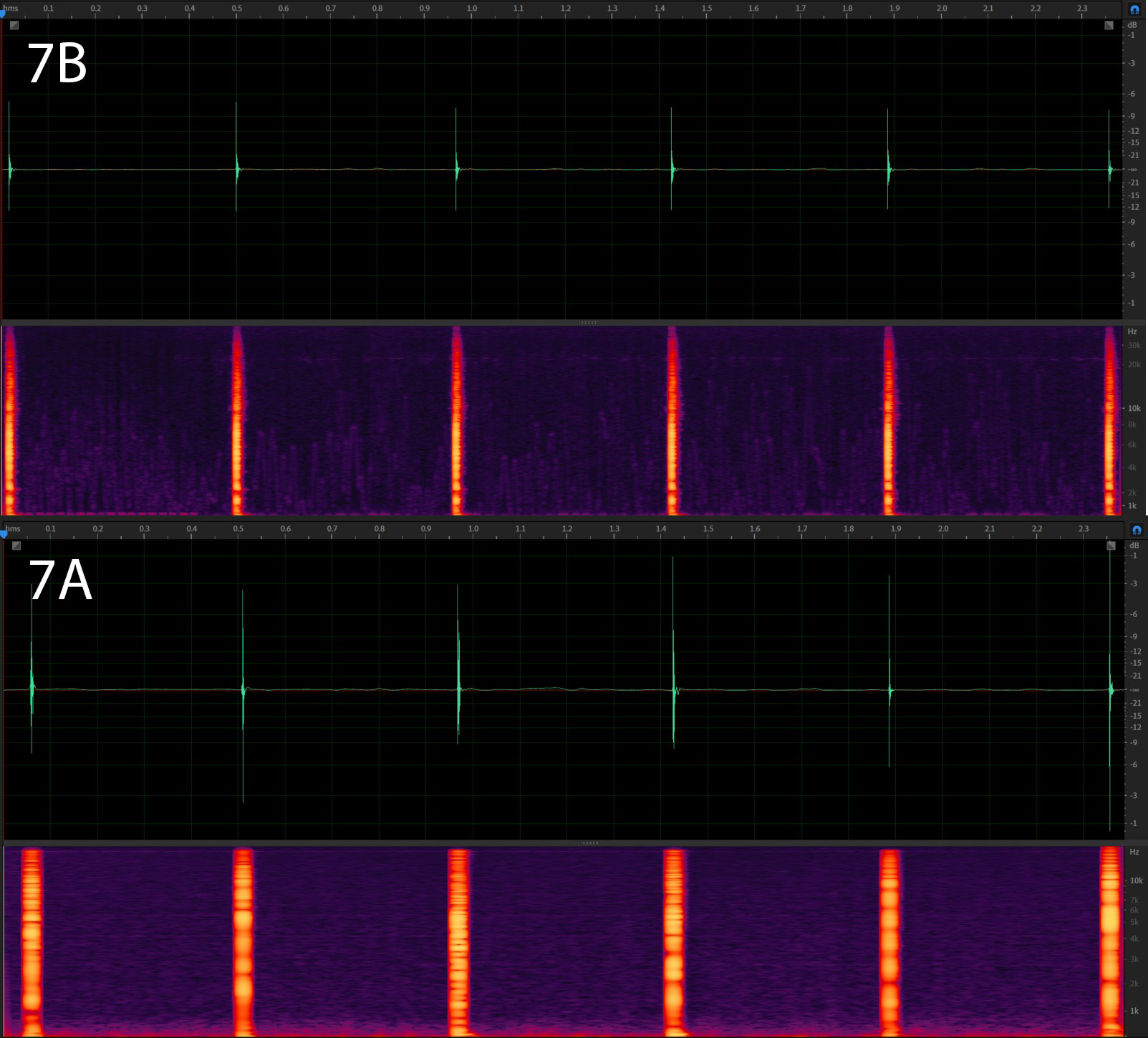}
    \caption{Waveform and spectrograms of both authentic clicks recorded from wild sperm whales from clip 7B (top) and from WhAM generated synthetic clicks from clip 7A (bottom). Here ``7'' is because this formed the seventh pair in the expert perceptual study (Part 1).}
    \label{fig:shane-fig4}
\end{figure*}

\newpage

\section{Additional experiments}

\subsection{FAD Embedding Selection}
\label{apx:fad_calibration}

The Fr'{e}chet Audio Distance (FAD) measures similarity between audio datasets using embeddings to map the audio into a feature space. The choice of embedding is crucial, as different embeddings capture different aspects of the signal. For analyzing sperm whale codas, we sought an embedding that prioritizes the temporal patterns critical to coda structure over background noise. This appendix describes the calibration experiment we conducted to select the most suitable embedding for our FAD analysis.

Let the coda recordings in DSWP+CETI be denoted by $\{x_1, ..., x_n\}$, we:
\begin{enumerate}
    \item Created denoised versions $\{\hat{x}_1, ..., \hat{x}_n\}$ as detailed in \Cref{apx:data}
    \item Isolated the removed noise components $\{x_1 - \hat{x}_1, ..., x_n - \hat{x}_n\}$
    \item For each candidate embedding $f_i$, compared:
    \begin{itemize}
        \item $d_1^i = $ FAD score between codas and their denoised versions:\\
        \item $d_2^i = $ FAD score between codas and their noise components:\\
        
    \end{itemize}
\end{enumerate}

We evaluated five audio embeddings VGGISH \cite{VGGish1,VGGish2}, Encodec-embd \cite{DefossezCSA23}, LAION CLAP Music, LAION CLAP Audio \cite{laionclap2023,htsatke2022}, and BirdNET \cite{BirdNet21} using the Fr\'{e}chet Audio Distance implementation of \citet{FADTK}. The ratio $d_2^i/d_1^i$ indicates how much more weight embedding $i$ gives to background noise versus temporal structure. A larger ratio indicates stronger emphasis on temporal patterns and better suitability for the quantitative assessment of audio translation experiment. \Cref{tab:fad_calibration} shows these ratios for each embedding.

\begin{table}[ht]
\caption{Comparison of Audio Embeddings for Temporal Structure Sensitivity}
\label{tab:fad_calibration}
\vskip 0.15in
\begin{center}
\begin{small}
\begin{sc}
\begin{tabular}{lccc}
\toprule
Embedding & $d_1$ (Coda vs. Denoised) & $d_2$ (Coda vs. Noise) & $\nicefrac{d_2}{d_1}$ \\
\midrule
VGGISH &  2.0844 & 1.5027 & 0.7209 \\
Encodec-embd & 25.9716 & 3.156 & 0.1215 \\
LAION CLAP Music & 0.1483 & 0.1080 & 0.7282 \\
LAION CLAP Audio & 0.1144 & 0.1098 & 0.9597\\
BirdNET & 16.7817 & 22.4761 & 1.3393\\
\bottomrule
\end{tabular}
\end{sc}
\end{small}
\end{center}
\vskip -0.1in
\end{table}

Based on these results, we selected BirdNET for our main FAD experiments, as it maximized the ratio of distances between raw-to-noise over raw-to-signal.

\subsection{Downstream Task Ablation Study}\label{apx:downstreamAblation}

To evaluate the contributions of different components in WhAM, we conduct an ablation study by progressively removing elements and assessing performance across the same set of downstream tasks. The results are presented in \Cref{fig:ablation}.

\paragraph{No finetuning.} We test the effect of skipping domain-adaptation (step (b) in \Cref{fig:main}), or skipping finetuning of VampNet altogether (steps b,c) in \Cref{fig:main}). For all tasks except Social Unit classification, removing species-specific finetuning or domain adaptation does not have a significant impact on the accuracy. This indicates that the inclusion of these steps in WhAM does not significantly degrade the performance on most downstream tasks.

\paragraph{Tokenizer-only.} We falsify the hypothesis that the neural audio codec is sufficient for capturing semantic properties in the audio by testing downstream classification directly on the acoustic tokens (\Cref{fig:vampnet}), without embedding them through the MATM. This causes a statistically significant performance drop, particularly in Social Unit classification (-10.9 points, from 70.5\% $\pm$ 0.7\% to 59.6\% $\pm$ 2.0\%)

\newcommand{\ablationfigwidth}{0.4\textwidth}
\begin{figure}[ht]
    \centering
    \begin{minipage}[b]{\ablationfigwidth}
        \centering
        \includegraphics[width=\textwidth]{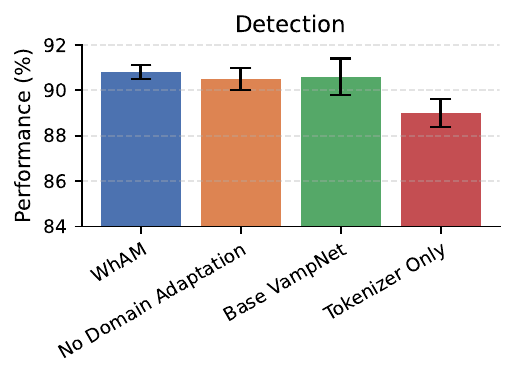}
    \end{minipage}
    \hspace{0.5cm}
    \begin{minipage}[b]{\ablationfigwidth}
        \centering
        \includegraphics[width=\textwidth]{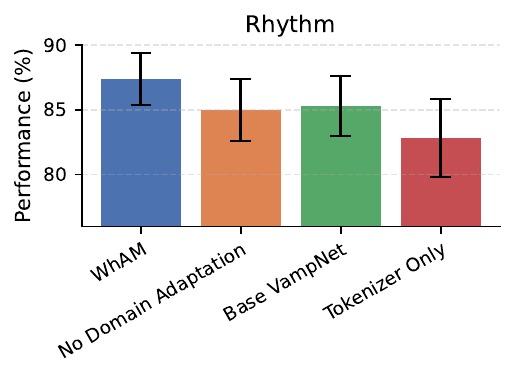}
    \end{minipage}
    
    \vspace{0.5cm}
    
    \begin{minipage}[b]{\ablationfigwidth}
        \centering
        \includegraphics[width=\textwidth]{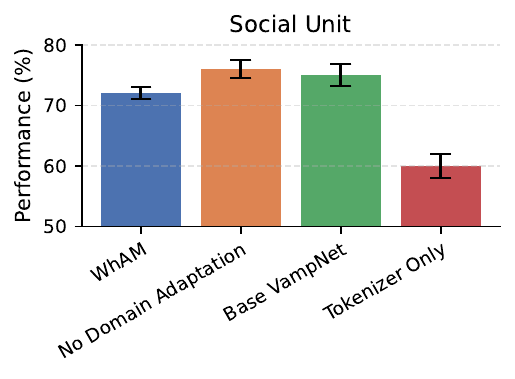}
    \end{minipage}
    \hspace{0.5cm}
    \begin{minipage}[b]{\ablationfigwidth}
        \centering
        \includegraphics[width=\textwidth]{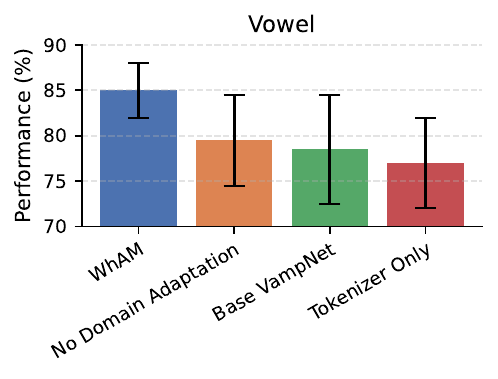}
    \end{minipage}
    \caption{Classification accuracies (\%) resulting from using the output of different WhAM components on downstream tasks. Each classifier head was trained using three different random seeds, with mean$\pm$stderr reported. }
    \label{fig:ablation}
\end{figure}

 \subsection{Fr\'{e}chet Audio Distance Ablation Study}\label{apx:fad_ablation} %
\begin{figure}
    \centering
    \includegraphics[width=1\linewidth]{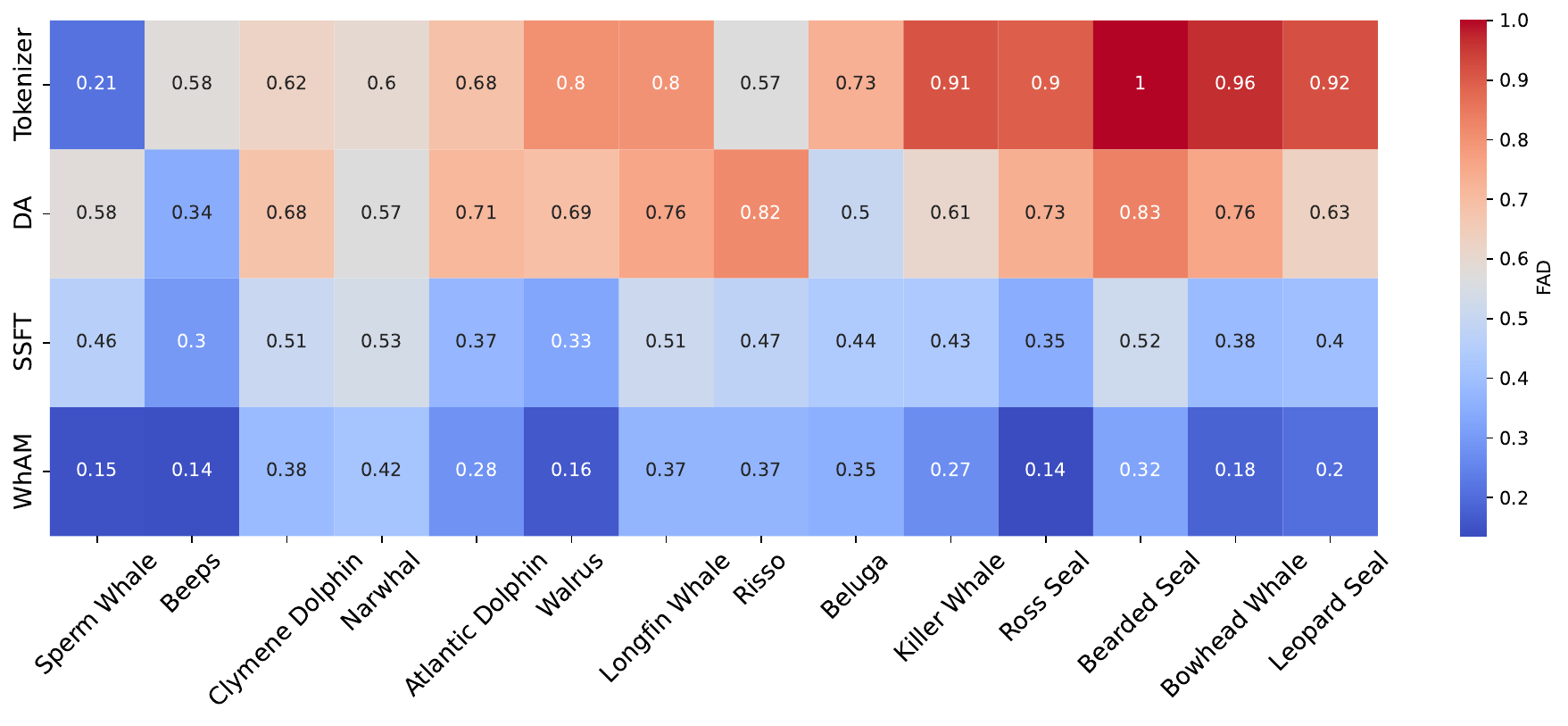}
    \caption{The effect of ablating components of the model on FAD}
    \label{fig:generative_ablations}
\end{figure}

To complement the ablation study of \Cref{apx:downstreamAblation}, the experiments detailed in \cref{sec:fad} were repeated four times with marine mammal sounds. First, using only the tokenizer. Second, training the model with only \textbf{Domain Adaptation} (DA, step (c) in \Cref{fig:main}), skipping the \textbf{Species Specific Fine-Tuning} step (SSFA, step (c) in \Cref{fig:main}). Third, training only with \textbf{Domain Adaptation}.  And finally, using the full version of WhAM. These results (\Cref{fig:generative_ablations}) show that, as expected, fine-tuning WhAM on sperm whale data results in outputs that are more similar to sperm whale vocalizations.

\subsection{Tokenizer Reconstruction Loss Study}

WhAM uses the Descript Audio Codec (DAC) as its tokenizer \citep{KumarSLKK23}.DAC  is tailored towards speech, music, and environmental sounds. To test possible degradation in encoding sperm whale coda audio, we conducted the following experiment.

Let each individual coda recording in the DSWP+CETI datasets be denoted by ${x_{1},x_{2}...x_{n}}$

\begin{enumerate}
    \item For each recording $x_{i}$, a reconstructed version, $\hat{x_{i}}$ was created by passing $x_{i}$ into the tokenizer to generate tokens, then passing the tokens through the decoder to recover the audio recording.
    \item Each $x_{i}$ and $\hat{x_{i}}$ was then sliced into chunks of length $C$ to calculate their respective short term fourier transforms. The transform is represented by the arrays  $\{ \Vec{x}_{i,1}, \Vec{x}_{i,2}...\Vec{x}_{i,m}\}$ and $\{ \Vec{\hat{x}}_{i,1}, \Vec{\hat{x}}_{i,2}...\Vec{\hat{x}}_{i,m}\}$. Each $\Vec{x}_{i,j}$ represents the magnitude all frequencies over the $j$th chunk of the recording $x_{i}$
    \item The mean reconstruction accuracy, denoted by $E$ is now given by taking the average normalized error between all $\Vec{x}_{i,j}$ and $\Vec{\hat{x}}_{i,j}$ using the formula: 
        $\frac{1}{n\times m}\sum_{i,j} (\Vec{x}_{i,j} - \Vec{\hat{x}}_{i,j})^{2}/(\Vec{x}_{i,j})^{2}$

\end{enumerate}    

\begin{figure}[ht]
    \centering
    \includegraphics[width=1\textwidth]{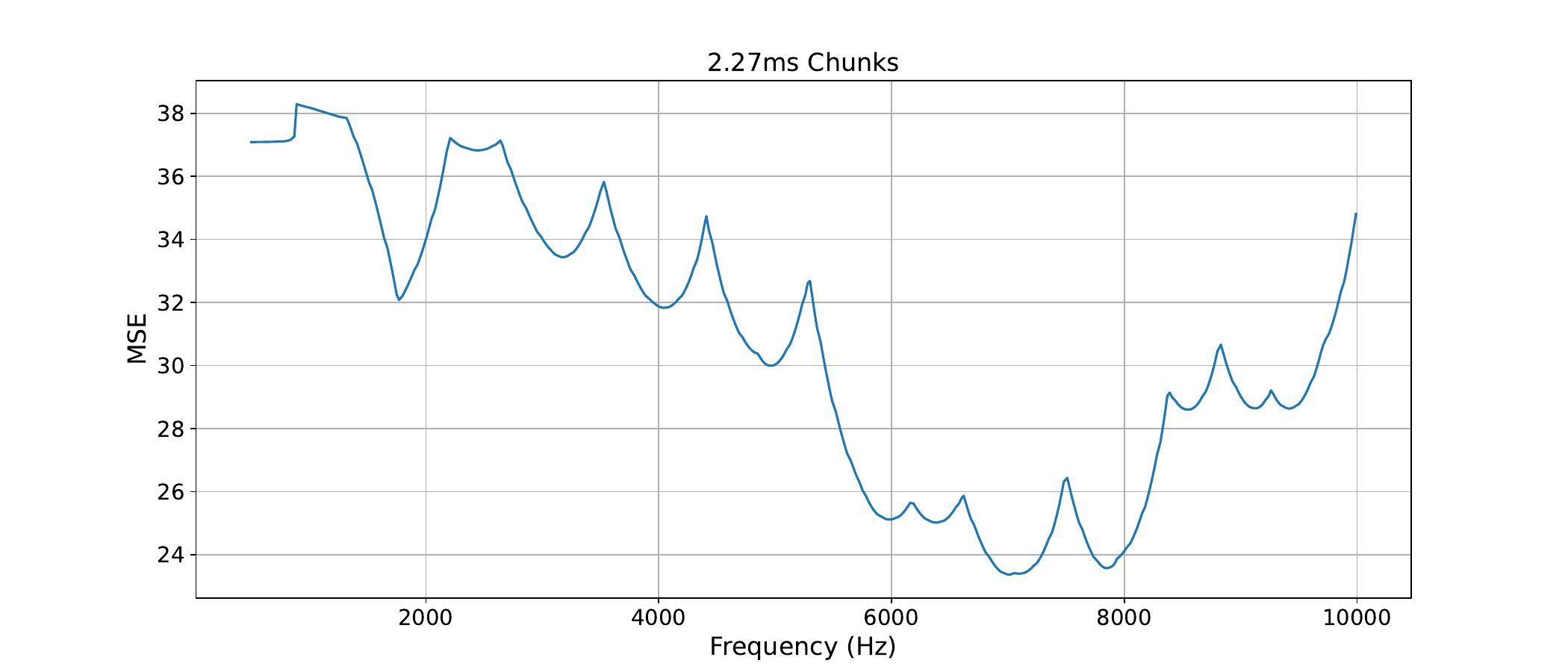}
    
    \vspace{0cm} %
    
    \includegraphics[width=1\textwidth]{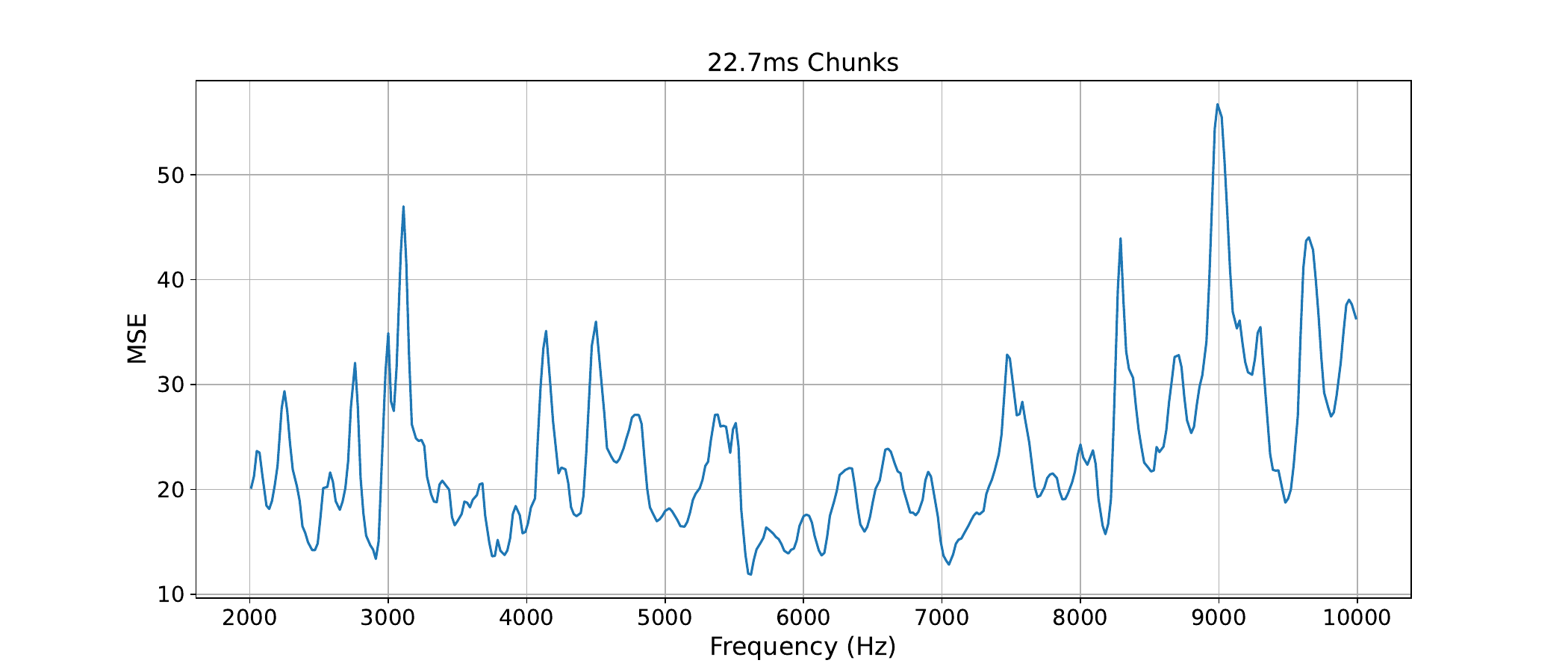}
    \caption{Tokenizer reconstruction loss study. Normalized mean squared error (y-axis) by frequency (x-axis).}
    
    \label{fig:fourier}
    
\end{figure}

The mean error is shown in \cref{fig:fourier} for a chunk size of 2.27 ms and 22.7 ms. With a smaller chunk size, $E$ indicates which pitches the tokenizer accurately reconstructs and which pitches it does not. Using a larger chunk size, $E$ gives an indication of what type of general noise patterns the tokenizer fails to include in its reconstruction. 

As can be seen with a chunk size of $22.7$ ms, the error is wave-like. Since sinusoidal waves in the frequency domain correspond to impulses in the time domain, this suggests a tendency to misrepresent impulse-like sounds in the time domain. On the other hand, using a chunk size of $2.27$ ms provides an indication as to what pitches the tokenizer prioritizes. The spikes in the bands 1--6 kHz and 8--10 kHz suggest that, in general, the tokenizer tends to perform relatively poorly in those frequencies. However, this degradation is not severe enough to prevent our model from generating natural-sounding codas (\Cref{sec:listening}), nor from its embeddings to ``capture'' vowels (\Cref{sec:downstream}). 

\newpage

\section{Methodology Details}\label{apx:methods}
\subsection{Data} \label{apx:data}

\paragraph{FSD.}
The FSD50k dataset includes 3,159 audio recordings labeled with the ``animal'' tag, amounting to a total duration of 7 hours and 45 minutes. Noisy segments were retained to preserve real-world variability in training data.

\paragraph{AudioSet.}
The AudioSet dataset was used to supplement training with additional animal vocalizations. It contains 5h8m hours of audio.

\paragraph{BirdSet.}  Consists of 6,800 total hours of recordings containing bird vocalizations \citep{Birdset}. Due to space constraints and to avoid training WhAM on audio that did not contain any vocalizations, only a subset of the entire dataset was used, containing a total of 110 hours of data.

\paragraph{WMMS.}
The Watkins Marine Mammal Sound Database consists of raw, unlabeled audio recordings. The dataset contains a total of 4 hours and 8 minutes of audio. Each recording was segmented into 10-second snippets for training. No additional denoising was applied. The dataset contained vocalizations from the following mammals (names as listed on the WMMS website): 

\begin{tasks}[after-item-skip=0.5em,label={}](3)
    \task Atlantic Spotted Dolphin
    \task Bearded Seal
    \task Beluga (White Whale)
    \task Bottlenose Dolphin
    \task Boutu (Amazon River Dolphin)
    \task Bowhead Whale
    \task Clymene Dolphin
    \task Commerson's Dolphin
    \task Common Dolphin
    \task Dall's Porpoise
    \task Dusky Dolphin
    \task False Killer Whale
    \task Fin, Finback Whale
    \task Finless Porpoise
    \task Fraser's Dolphin
    \task Grampus (Risso's Dolphin)
    \task Gray Seal
    \task Gray Whale
    \task Harbor Porpoise
    \task Harbour Seal
    \task Harp Seal
    \task Heaviside's Dolphin
    \task Hooded Seal
    \task Humpback Whale
    \task Irrawaddy Dolphin
    \task Juan Fernandez Fur Seal
    \task Killer Whale
    \task Leopard Seal
    \task Long-Beaked (Pacific) Common Dolphin
    \task Long-Finned Pilot Whale
    \task Melon-Headed Whale
    \task Minke Whale
    \task Narwhal
    \task New Zealand Fur Seal
    \task Northern Right Whale
    \task Pantropical Spotted Dolphin
    \task Ribbon Seal
    \task Ringed Seal
    \task Ross Seal
    \task Rough-Toothed Dolphin
    \task Sea Otter
    \task Short-Finned (Pacific) Pilot Whale
    \task Southern Right Whale
    \task Sperm Whale
    \task Spinner Dolphin
    \task Spotted Seal
    \task Steller Sea Lion
    \task Striped Dolphin
    \task Tucuxi Dolphin
    \task Walrus
    \task Weddell Seal
    \task West Indian Manatee
    \task White-beaked Dolphin
    \task White-sided Dolphin
\end{tasks}

\paragraph{DSWP.} The dataset consists of codas collected between 2005--2018 in a 2000km$^2$ area off the coast of Dominica. Codas were recorded using various recording systems including far-field boat-based hydrophones and animal-borne tags. Recording setups were as follows:
\begin{description}
	\item[2005:] A Fostex VF-160 multitrack recorder (44.1kHz sampling rate) and a custom built towed hydrophone (Benthos AQ-4 elements, frequency response: 0.1--30kHz) with a filter box with high-pass filters up to 1 kHz resulting in a recording chain with a flat frequency response across a minimum of 2--20kHz.
	\item[2006:] No recordings during this short season.
	\item[2007,2009,2011:] A Zoom H4 portable field recorder (48kHz sampling rate) and a Cetacean Research Technology C55 hydrophone (frequency response: 0.02--44kHz) and no filters. 
	\item[2008,2010,2012,2015:] A custom-built towed hydrophone (Benthos AQ-4 elements, frequency response: 0.1--30kHz) with a filter box with high-pass filters up to 1 kHz resulting in a recording chain with a flat frequency response across a minimum of 2--20 kHz. This was connected to a computer based recording system as a part of the International Fund for Animal Welfare’s (IFAW) LOGGER software package (48kHz sampling rate) or PAMGUARD (minimum 48 kHz sampling rate). In addition, recordings were also made through the deployment of animal-borne sound and movement tags (DTag generation 3, \citealt{JohnsonT03}).
\end{description}
\paragraph{CETI.} All systems were sampling above 96kHz with a 16bit resolution with a minimum flat ($\pm$2dB) frequency response within 1--45kHz.

The DSWP and CETI dataset contain background noise such as water sounds. To improve model performance, we denoise datasets before training on the model. A noise profile of each recording in the frequency domain was generated by sampling sections which did not contain codas. Then, we perform spectral subtraction to remove noise in the frequency domain, and transform back to the time domain of the audio signal.

All audio samples were downsampled to 16 kHz and normalized to have zero mean and unit variance when passed into VampNet.

Note that, as with any self-supervised training setup that relies on random masking, the effective number of unique training examples far exceeds the raw audio hours. In our case: First, each 2-second audio snippet becomes a $14 \times 120$ token array. Columns correspond to time steps, and rows represent acoustic granularity During training, entire columns (i.e., time steps) are masked at random; with 120 columns, there are $2^{120}$ possible masking patterns per snippet. So, for example a 20 hour dataset yields 36,000 snippets, which result in $\approx 10^{40}$ possible masked training inputs.

\subsubsection{\texorpdfstring{Generating data for \Cref{sec:fad,sec:listening}}{Generating data}}

Three different input sources were used to generate samples for both the \textbf{Quantitative Assessment of Audio Translation} and the \textbf{Expert Perceptual Evaluation}. The prompt settings for each input type are summarized in \Cref{tab:prompt_settings}.

\begin{table}[t]
\caption{Quantitative Assessment Data Summary}
\label{tab:marine_animals}
\vskip 0.15in
\begin{center}
\begin{small}
\begin{sc}
\begin{tabular}{lcc}
\toprule
Full Name & Shortened Name & Num. Samples \\
\midrule
Atlantic Dolphin & A. Dolphin & 58 \\
Bearded Seal & B. Seal & 37 \\
Bowhead Whale & B. Whale &  60\\
Beluga Whale, White Whale & Beluga & 50\\
Walrus & Walrus &  38\\
Clymene Dolphin & C. Dolphin & 63 \\
Narwhal & Narwhal &  50\\
Leopard Seal & L. Seal & 10 \\
Long-finned Whale & L. Whale & 10 \\
Killer Whale (Orca) & Orca &  35 \\
Ross Seal & Ross Seal & 50 \\
Risso's Dolphin & Risso & 67 \\
\bottomrule
\end{tabular}
\end{sc}
\end{small}
\end{center}
\vskip -0.1in
\end{table}

\begin{table}[t]
\caption{Prompt settings for each input type.}
\label{tab:prompt_settings}
\vskip 0.15in
\begin{center}
\begin{small}
\begin{sc}
\begin{tabular}{lccccc}
\toprule
Input & Periodic  & Onset & Num. of & Typical  & Sample \\
 & Prompt &  Mask Width& Steps & Mass & Cutoff \\
\midrule
Codas & 12 & 21 & 50 & 0.102 & 0.17 \\
Beeps & 12 & 21 & 50 & 0.102 & 0.17 \\
A. Dolphin & 16 & 5 & 74 & 0.15 & 0.39\\
B. Seal & 7 & 1 & 70 & 0.15 & 0.44\\
B. Whale & 7 & 1 & 70 & 0.15 & 0.44\\
Beluga & 13 & 13 & 85 & 0.15 & 0.39\\
Walrus & 18 & 1 & 107 & 0.15 & 0.33 \\
C.Dolphine & 12 & 14 & 72 & 0.15 & 0.25  \\
Narwhal & 6 & 4 & 39 & 0.15 & 0.21  \\
L. Seal & 6 & 4 & 46 & 0.15 & 0.39  \\
L. Whale & 15 & 19 & 57 & 0.15 & 0.42  \\
Orca & 13 & 2 & 46 & 0.15 & 0.39 \\
Ross Seal & 18 & 3 & 66 & 0.15 & 0.49 \\
Risso & 13 & 13 & 85 & 0.15 & 0.39 \\
\bottomrule
\end{tabular}
\end{sc}
\end{small}
\end{center}
\vskip -0.1in
\end{table}
\paragraph{Watkins Marine Mammals.}
Eleven species were selected from the ``Best of Watkins Marine Mammals'' dataset. Due to variations in vocalization characteristics and recording conditions, prompt settings were manually optimized for each species. These species and prompt settings can be found in \Cref{tab:prompt_settings}.

\paragraph{Digital ``beeps''.}
Five digital beep sequences were generated. Each snippet was initialized as a zero-filled array at a 44.1 kHz sample rate. Clicks were simulated by selecting random indices and setting them to a peak amplitude of 1. To ensure realistic timing and rhythm, real coda sequences were prepended to each generated sample before synthesis. These prepended codas were then removed after generation.

\subsection{Model Training}\label{apx:training}
The model training procedure consisted of two phases: domain adaptation and species-specific fine-tuning. 

\paragraph{Acoustic Tokenizer Settings.} Discrete token vocabulary size ($\Sigma$) = 1024. Frequency of Input Audio $\Nsamples$ = $16$kHz. Tokenizer input length $\Nsec$ = 10.

\paragraph{Domain Adaptation.}
In the first phase, the model was pretrained on a mixture of general animal vocalizations, including data from FSD and AudioSet. This step aimed to establish a broad understanding of bioacoustic patterns. The model was trained for 500,000 iterations using the AdamW optimizer with a learning rate of 0.0001. A batch size of 6 was used, and gradient clipping was applied to stabilize training. The model took 123 hours to train using an AWS EC2 g5.2xlarge instance (NVIDIA A10 GPU, 8 vCPUs, 32 GB of memory).

\paragraph{Species-Specific Fine-Tuning.}
Following domain adaptation, the model was fine-tuned on whale-specific data from DSWP+CETI to adapt its representations to sperm whale vocalizations. The fine-tuning process used the same optimizer and learning rate as the pretraining phase and a batch size of 6. Training continued for another 500,000 iterations. This took 39 hours to run using an AWS EC2 g5.2xlarge.

\subsection{Computational costs}\label{apx:experiments}
All experiments were run on an AWS EC2 g5.2xlarge (NVIDIA A10 GPU, 8 vCPUs, 32 GB of memory). A full run of FAD experiments took 3 hours with a full version of Vampnet, and 1.5 hours using a Tokenizer-only model, therefore \Cref{sec:fad,apx:fad_ablation} took approximately 7.5 hours in total. \Cref{apx:fad_calibration} took approximately 1.5 hours. For downstream classification, training the linear probe took at most 5.5 hours; thus, \Cref{sec:downstream,apx:downstreamAblation} took about 16.5 hours in total.

\subsection{Utility of Embeddings for Downstream Tasks}\label{apx:downstream}

\paragraph{Model Details.} 
We run a forward pass through WhAM and AVES to obtain embeddings from the audio. Both WhAM and AVES output varying embeddings over time, so we average the embeddings over time to obtain 1 unified embedding for 1 audio snippet. After the embedding is obtained, we attach a two-layer feed-forward neural network as a classifier. The network consists of a fully connected layer that projects the embedding into a 128-dimensional hidden layer, followed by a ReLU activation. A second fully connected layer then generates class probabilities.

We evaluate embeddings from WhAM and AVES, comparing their performance against a random embedding baseline as well as a majority baseline classifier.

\paragraph{Training Data.} 
For downstream task evaluation, we leveraged annotations in the DSWP+CETI datasets. Using human-annotated timestamps, we identified and extracted audio segments containing codas, each spanning 1--2 seconds. Each coda was labeled for one of the following classification tasks:
\begin{itemize}
    \item \textbf{Coda Detection}: Determine whether a given audio snippet contains a whale coda.
    \item \textbf{Rhythm Type Classification}: Classify codas according to their rhythmic patterns. For this task, we choose to include samples whose rhythm types are among the 5 most common, because the remaining ones appear too infrequently for classifiers to be accurate.
    \item \textbf{Social Unit Classification}: Identify the social unit associated with each coda.
    \item \textbf{Vowel Classification}: Detect vowel-like patterns within whale vocalizations.
\end{itemize}

\Cref{tab:downstream_data} summarizes dataset sizes for each task.

\begin{table}[t]
\caption{Dataset sizes for downstream classification tasks.}
\label{tab:downstream_data}
\vskip 0.15in
\begin{center}
\begin{small}
\begin{sc}
\begin{tabular}{l c}
\toprule
\textbf{Task} & \textbf{Number of Samples} \\
\midrule
Coda Detection & 3,100 \\
Rhythm Type Classification & 916 \\
Social Unit Classification & 2,659 \\
Vowel Classification & 486 \\
\bottomrule
\end{tabular}
\end{sc}
\end{small}
\end{center}
\vskip -0.1in
\end{table}

\paragraph{Training Process.} 
We split the dataset into 80\% training and 20\% testing, using stratified sampling of labels to ensure consistent label distribution. The embedding model is frozen, and only the classifier parameters are trained. Training is performed on an NVIDIA A10G GPU for 10 epochs, using a learning rate of $10^{-4}$ and a batch size of 32. Model checkpoints are saved at each epoch, and the best-performing model is selected based on test set performance.

\subsection{Expert Perceptual Evaluation}\label{apx:listening}

Five domain experts in sperm whale bioacoustics participated in the evaluation. Given the highly specialized nature of sperm whale vocalization analysis, the pool of qualified experts with years of direct experience analyzing and annotating these vocalizations is notably small. All participants were recruited from an established research collaboration studying cetacean communication, and each had at least three years of experience working with sperm whale codas. No compensation was given to participants.

The evaluation was conducted via Google Form. The form began with the following introduction:

\begin{tcolorbox}[title=Welcome,colback=white,colframe=black]
Thank you for participating in this study. Your expertise in analyzing sperm whale vocalizations is invaluable for evaluating our model.

The study consists of four parts, to be completed in order. A final section includes three short questions about your background.

\textbf{Technical Setup}
\begin{itemize}
    \item Download and extract the \texttt{listener\_evaluation.zip} file from a provided link
    \item Use headphones for all listening tasks
    \item Complete the experiment in a quiet environment
    \item You can take breaks between sections as needed
\end{itemize}

If you encounter any technical difficulties or have questions about the procedure, please contact [omitted].

\medskip
\textbf{Participant Identification}\\
Name (used for tracking responses only): \rule{5cm}{0.1pt}
\end{tcolorbox}

\subsubsection{Audio-Only Two-Alternative Forced Choice (2AFC)}
Listeners were presented with 30 pairs of codas. Each pair contained an original, denoised coda and a model-generated counterpart. Participants were asked to identify which sample was the original and which was generated.

\begin{tcolorbox}[title=Task Instructions,colback=white,colframe=black]
In this section, you will listen to pairs of codas. For each pair, one is a natural recording and one is synthetic. Please indicate which one you believe is synthetic.

The audio files are located in the ***section1/*** folder. Each pair consists of two files:
\begin{itemize}
    \item *1a.wav* + *1b.wav*
    \item *2a.wav* + *2b.wav*
    \item etc.
\end{itemize}

Please listen to each file **at most three times**. Base your decision only on the provided audio. Do not visualize the audio.
\end{tcolorbox}

\subsubsection{Mixed Two-Alternative Forced Choice (2AFC)}
Listeners were presented with 25 individual samples: 10 real codas, 5 generated from real codas, 5 generated from walrus vocalizations, and 5 generated from digital beeps.
Each listener classified each sample as either real or generated.

\begin{tcolorbox}[title=Task Instructions,colback=white,colframe=black]
In this section, you will listen to individual codas and classify each as either natural or synthetic.

The audio files are located in the ***section2*** folder:
\begin{itemize}
    \item *1.wav*
    \item *2.wav*
    \item etc.
\end{itemize}

Please listen to each file at most three times. **Base your decision only on the provided audio. Do not visualize the audio.**
\end{tcolorbox}

\subsubsection{Visualized Two-Alternative Forced Choice (2AFC)}
This experiment was identical to the \textbf{Audio-Only 2AFC} condition, except participants were allowed to inspect the spectrograms of each recording using their preferred software before making their decision. Marine biologists preferred Adobe Auditions, while underwater acoustics experts used Matlab.

\begin{tcolorbox}[title=Task Instructions,colback=white,colframe=black]
Once again, you will listen to pairs of codas (a.wav and b.wav). For each pair, one is a natural recording and one is synthetic. Please indicate which one you believe is synthetic.

The audio files are located in the ***section3/*** folder. Each pair consists of two files:
\begin{itemize}
   \item *1a.wav* + *1b.wav*
   \item *2a.wav* + *2b.wav*
   \item etc.
\end{itemize}

Please listen to each file at most three times. **You may now visualize the audio using any software you are familiar with.**

What software will you use to visualize the audio?
\rule{5cm}{0.1pt}
\end{tcolorbox}

\subsubsection{Qualitative Assessment}

\begin{tcolorbox}[title=Task Instructions,colback=white,colframe=black]
For this final section, please first listen to the reference synthetic codas provided in the section4 folder. These examples were chosen to represent typical outputs of our model. Then, based on these examples and your experience with all parts of the experiment, please answer the following questions

What characteristics of natural codas are well represented in the synthetic ones?

What characteristics of natural codas are missing or different in the synthetic ones?

Did you observe any patterns in the synthetic codas that do not occur in natural ones?

When **only listening** to the audio (sections 1 and 2), what helped you distinguish between natural and synthetic codas?

When **visualizing** the audio pairs (section 3), what helped you distinguish between natural and synthetic codas?
\end{tcolorbox}

\subsubsection{Background Information}

\begin{tcolorbox}[title=Task Instructions,colback=white,colframe=black]
To help contextualize the evaluations, please tell us about your experience working with sperm whale codas.

How many years have you spent professionally analyzing sperm whale codas (e.g., in research, conservation, or educational contexts)? 

What types of coda work have you performed?
\begin{itemize}
   \item *Recording of codas in the field*
   \item *Development of recording methods for codas*
   \item *Manual detection, classification or annotation of codas*
   \item *Development of automatic detection, classification or annotation systems*
   \item *Meta-analysis (e.g. methodology development, literature review)*
   \item *Other...*
\end{itemize}

In what contexts have you worked with coda recordings?
\begin{itemize}
   \item *Academic research*
   \item *Conservation work*
   \item *Industry/commercial projects*
   \item *Educational/training contexts*
   \item *Government/regulatory work*
\end{itemize}

What is your primary field of expertise?
\end{tcolorbox}
\end{document}